\documentclass[11pt]{article}
\usepackage[margin=1in]{geometry}
\usepackage[utf8]{inputenc}
\usepackage[T1]{fontenc}
\usepackage{hyperref,graphicx}
\usepackage{url}
\usepackage{booktabs}
\usepackage{amsfonts}
\usepackage{nicefrac}
\usepackage{bbm}
\usepackage{microtype}
\usepackage{xcolor}
\usepackage[numbers,sort&compress]{natbib}
\usepackage{algorithm}
\usepackage{amsthm,amsmath}
\usepackage{algpseudocode}
\newtheorem{theorem}{Theorem}[section]

\newtheorem{proposition}[theorem]{Proposition}
\newtheorem{assumption}[theorem]{Assumption}

\title{Flow-Controlled Scheduling for LLM Inference with Provable Stability Guarantees}

\date{}

\author{%
  Zhuolun~Dong\thanks{Department of Information, Risk, and Operations Management, 
  The University of Texas at Austin. Email: \texttt{dongzhuolun@utexas.edu}.} \and
  Junyu~Cao\thanks{Corresponding author. Department of Information, Risk, and 
  Operations Management, The University of Texas at Austin. 
  Email: \texttt{junyu.cao@mccombs.utexas.edu}.}
}

\begin{document}

\maketitle

\begin{abstract}
Large language models (LLMs) have been widely adopted due to their great performance across a wide range of applications. ChatGPT and Gemini now serve hundreds of millions of active users and handle billions of user requests per day, which puts optimizing LLM inference into the spotlight. A key challenge in LLM inference is that decode lengths are unknown. The memory usage for each request grows with generated tokens, which may lead to overflow and cause system instability. To address this concern, we propose a simple flow-control framework that controls the rate at which prompts join the active set. We derive a necessary condition that any stable system must satisfy and establish sufficient conditions under which our algorithm provably achieves stability. Experiments show that, compared to commonly used strategies in practice, our approach achieves higher token and request throughput, lower average and tail latency, and more stable KV cache utilization.
\end{abstract}

\section{Introduction}
Large Language Models (LLMs) have become a core component of modern AI applications, enabling a wide range of capabilities, from creative writing and translation to code generation and reasoning. Built on the transformer architecture \citep{vaswani2017attention}, these models have rapidly scaled in both size and performance. For instance, Llama~3 \citep{grattafiori2024llama} has 405 B parameters, and DeepSeek \citep{liu2024deepseek} has 701 B parameters. This rapid growth has brought unprecedented serving demand: \citet{chatterji2025people} documented that, by July 2025, 700 million users were sending 18 billion requests per week on ChatGPT. Therefore, the efficiency and reliability of LLM inference (serving) are critical bottlenecks for both cost and user experience. Characterized by its autoregressive generation behavior, LLM inference systems process customer requests in two distinct stages: a \emph{prefill} stage that processes the input prompt, followed by a \emph{decode} stage that generates output tokens sequentially. This complex serving mechanism, combined with latency and throughput requirements, poses challenges for inference systems.

To improve GPU utilization, modern inference systems apply a batch policy, where requests are batched and handled simultaneously. Under this batching paradigm, a central challenge is the efficient management of the GPU memory, specifically the Key-Value (KV) cache. During inference, the system caches the key and value computed from the previously generated prefill and decode tokens to generate the subsequent tokens. The memory footprint of a request grows accordingly during token generation and is released only when the request is completed, introducing dynamic memory consumption. In addition, real-world request arrivals are highly stochastic, with variable input and output lengths. Therefore, the aggregate memory usage under the batching policy may exhibit significant fluctuation. When memory usage approaches the limit, the system may be forced to preempt or drop requests, which degrades the latency performance and user experience. Further, the system needs to make instantaneous decisions, which introduces additional challenges.

To mitigate these challenges, researchers in computer systems have proposed inference engines with continuous batching scheduling mechanisms (e.g., Orca, vLLM) that allow token-level batching \citep{yu2022orca,kwon2023efficient}. These strategies prioritize the prefill tokens for incoming requests to maximize throughput and reduce latency. Subsequently, \citep{agrawal2023sarathi} employed chunked-prefills, splitting a prefill request into chunks of equal size. This approach efficiently prevents head-of-line blocking due to requests with long prefill lengths and results in good empirical performance. 

While recent LLM serving systems have substantially improved throughput and latency, their scheduling logic is often motivated by heuristic intuition. A theoretical framework for assessing the performance of scheduling algorithms under varying workload and resource conditions remains underdeveloped. Meanwhile, the operations research community has developed a rich body of theory and algorithms for resource allocation and scheduling problems, with impactful applications in cloud and distributed systems \citep{mitzenmacher2002power,lu2011join,verma2015large,banerjee2025load}, service operation systems \citep{gurvich2009scheduling,bansal2001analysis,iyer2023achieving,guo2023signaling}, and healthcare \citep{hu2022optimal,chen2025optimal,bauerhenne2026robust}. LLM inference systems form a new class of scheduling problems in which request arrivals are stochastic with heterogeneous process times. These features create opportunities for applying queueing analysis, as argued by \citep{mitzenmacher2025queueing}. Currently, LLM inference studies driven by operations research techniques remain in their early stages. Examples include \citep{jaillet2025online,li2025throughput,ao2025optimizing,wang2025llm}.  

Motivated by this gap, we study LLM inference through the lens of queueing theory. We model the inference engine as a stochastic service system in which the GPU memory is the main constraint. Our key insight is to focus on request admission (i.e., when a request enters the active set and begins to be prefilling). When many requests are processed simultaneously, their memory usage tends to increase simultaneously, leading to concentrated memory peaks. Under a high workload, this may trigger preemption or dropping, thereby affecting the system performance.  

Building on this insight, we propose a simple scheduling strategy, where we control the rate at which requests are activated. By doing so, we aim to avoid bursty admissions, which allows the memory usage to vary more smoothly. We develop a queueing model of LLM inference and derive a necessary condition that any stable system must satisfy (i.e., when violated, latency explodes over time) under any scheduling policy. Meanwhile, we establish sufficient conditions for stability and characterize an easy-to-implement algorithm to achieve it. Our contributions are as follows.

\begin{itemize}
    \item We propose a queueing-theoretic model to analyze the operations of LLM inference systems. In particular, we discretize the LLM serving system to the token level. At each time step, the decision-maker can assign a batch of active requests and generate an additional token for them. This model transforms a complex inference system into a queueing model, enabling us to derive useful technical results.
    \item Our theoretical analysis reveals a necessary condition that any stable system must satisfy: when the expected workload exceeds the KV cache capacity, no scheduling algorithm can prevent latency from growing unboundedly.
    \item We establish a new scheduling algorithm that activates new requests smoothly. We show that our algorithm can be stable under a particular load condition. These results offer practical implications for LLM inference.
    \item We conduct comprehensive experiments using both synthetic and real-world datasets from \citep{zheng2023lmsys}. Our method has good empirical performance across both high-load and low-load conditions. In particular, our scheduling algorithm outperforms benchmark algorithms in both token throughput and request completion rate.
\end{itemize}

\subsection{Related Work}

\paragraph{LLM Inference.} Modern LLM inference engines improve throughput and tail latency under tight GPU memory constraints. Early systems such as FasterTransformer \citep{fast} allow request-level scheduling while prioritizing the decoding stage. Recent engines enable continuous batching that prioritizes the prefill stage to maximize token throughput \citep{yu2022orca}. \citep{kwon2023efficient} proposes PagedAttention, which divides requests' KV cache into blocks to improve concurrency. Subsequent work further improved the KV usage by employing chunked prefills to reduce head-of-line blocking \citep{agrawal2023sarathi}. These engines are proved to have good empirical evidence in practical workloads. Moreover, these works colocate both stages on the same GPU and batch the computation of prefill and decoding across all requests. By contrast, some other engines (e.g., \cite{patel2024splitwise,zhong2024distserve}) split the two phases on separate machines and achieve good throughput with lower costs.

\paragraph{Queueing Theory.} The queueing community has established an immense body of works on studying the behavior of stochastic service systems under resource constraints. This line of literature characterizes when such systems can operate stably and analyzes the system performance (e.g., throughput or delay) with multiple request types \citep{chen2001fundamentals,hu2022optimal,chen2025optimal}. These results are translated and applied into broad application domains, including service systems \citep{guo2023signaling,zhong2025learning}, healthcare \citep{hu2022optimal}, and cloud systems \citep{lu2011join}. 

LLM inference systems are related to stochastic service systems because the customer requests arrive stochastically, with heterogeneous input and output lengths. Due to the complex nature of inference systems, they offer unique challenges to the queueing community, as highlighted by \citep{mitzenmacher2025queueing}. Recent works develop formalized models for the LLM inference and establish theoretical guarantees \citep{ao2025optimizing,li2025throughput,jaillet2025online,wang2025llm}. \citep{li2025throughput} proves that a class of work-conserving scheduling algorithms are throughput-optimal. \citep{ao2025optimizing} proposes a new batch policy with provable guarantees. Numerical experiments suggest that their algorithm achieves competitive performance with the state-of-the-art schedulers. \citep{jaillet2025online,wang2025llm} propose polynomial-time algorithms that are shown to be robust. Complementary to these efforts, our work focuses on mitigating system instability under variable prefill and decode lengths.

\section{Model and Problem Formulation}

\subsection{Problem Formulation}
We study an online LLM inference problem on a single computational worker (e.g., GPU) with a limited key-value (KV) cache budget $M$. This budget represents the maximum amount of GPU memory that can be used to store KV cache states, which depends on the model and the hardware. We consider a discrete-time framework. Time is slotted and indexed by $t \in \mathbb{N}$. Each request (prompt) $p_i$ is characterized by a triplet $(l_i, o_i, t_i)$, where $l_i \in \mathbb{N}$ denotes the prompt length (in tokens), $o_i \in \mathbb{N}$ denotes the length of output tokens to be generated, and $t_i \in \mathbb{N}$ denotes the arrival time. The decision-maker observes a request only when it arrives. Upon arrival at time $t_i$, the prompt length $l_i$ is revealed, while the output length $o_i$ may not be known in advance. 

At the beginning of each time step $t$, a collection of new prompts arrives (denoted by $\mathcal{A}_t$). In particular, $n_t = |\mathcal{A}_t|$ is sampled from a known distribution $\mathbb P(\cdot)$ with mean $\lambda$. These newly arrived requests first join a waiting queue, denoted by $\mathcal{Q}_t$, waiting to be activated. For any request $p_i$, the decision-maker chooses a starting time $t_i'\in \mathbb N$ ($t_i' \ge t_i$). Once activated, the request joins the active set and begins to admit to KV cache memory. A request remains in the active set until it either completes generating all output tokens or is evicted. Once a request is evicted, it is refreshed and will return to the waiting queue. Define the active set at $t$ as $\mathcal{P}_t$. For each active request $p_i \in \mathcal{P}_t$, let $j_{i(t)} \in \{0,1,\dots,o_i\}$ denote the number of output tokens that have been generated at the end of time slot $t$. The decision-maker specifies a subset $\mathcal{S}_t\subseteq \mathcal{P}_t$ to decode, where each request in $\mathcal{S}_t$ generates exactly one additional output token during the time slot. Therefore, the memory usage of the prompt $p_i$ at time $t$ is $l_i + j_{i(t)}$. Note that the overall KV cache usage cannot exceed the limit $M$, we have
\[
U_t = \sum_{p_i \in \mathcal{P}_t} (l_i + j_{i(t)}) \leq M, \quad \forall t.
\]
When the request $p_i$ is generating its last token (i.e., $j_{i(t)} = o_i$), the memory footprint reaches $l_i + o_i$ during that slot, and the request then completes and immediately releases its KV cache at the end of the same slot.

\subsection{Evaluation Metrics}\label{sec2.2}
We now define evaluation metrics along two complementary dimensions: \emph{efficiency}, and \emph{responsiveness}. These metrics measure the efficiency and the responsiveness of the LLM inference systems.

\paragraph{Throughput.} Throughput captures the system's ability of prefilling and decoding, quantifying the system efficiency. We consider two notions of throughput:

\begin{itemize}
    \item \emph{Token Throughput:} The number of tokens generated per unit time.
    \item \emph{Prompt Throughput:} The number of prompts completed per unit time.
\end{itemize}
These metrics capture different aspects of the system performance and are not equivalent. A policy that prioritizes requests with longer prefill or decode lengths may sustain high token throughput but perform poorly in prompt throughput because fewer requests are completed per unit of time.

\paragraph{Latency.}
The second metric is the end-to-end latency, equivalent to the time from arrival to completion. Latency accounts for both waiting time (before activation) and the decoding time, measuring the responsiveness of the system.

In this work, we use these metrics to evaluate the system more comprehensively. Throughput reflects the system’s efficiency in utilizing the GPU to generate tokens and complete prompts. Latency captures the responsiveness of the service from the user’s perspective.

\section{Methodology}
In the previous section, we introduce an LLM inference framework and evaluation metrics. In this section, we now present a unified inference approach that controls the rate at which incoming requests are admitted into the active set. This \emph{flow-control} design resolves a trade-off between throughput and system stability under memory constraints while incurring only small impact on latency. We also develop theoretical guarantees for the proposed approach.

\subsection{Flow-Control for LLM Inference}
The core principle of our flow-control design is to control the activation rate. Specifically, at each time step, we impose an activation budget and allow only a limited number of prompts to join the active set. This mechanism prevents transient surges in memory usage, which could inflate KV cache usage and destabilize the system. Moreover, the flow-control design also smooths the memory usage: when the system admits many prompts in a short interval, it may follow a burst of simultaneous completions, leading to a drop of memory and causing volatility. In contrast, controlling the activation rate stabilizes the inference system, yields a more predictable KV cache usage trajectory, and improves stability.

\subsection{Known Output Length}
We begin with a simplified setting, where the decision-maker observes both the prefill and decode lengths of a request upon arrival. While the autoregressive nature of LLM response generation implies stochastic decode lengths, recent studies \citep{zheng2023response} have shown significant progress in output length prediction. This assumption is also adopted in \citep{jaillet2025online}. By analyzing this scenario, we are able to derive sharper results. In the following Section \ref{sec3.3}, we will relax this assumption.

We classify requests into $m$ classes according to their prefill and decode lengths. For each class $k \in [m] = \{1,2\ldots,m\}$, requests share the same prefill length $l^{(k)}$ and decode length $o^{(k)}$. This is practically meaningful because real-world requests typically exhibit input and output lengths with hundreds of tokens \citep{zheng2023lmsys} despite their heterogeneity. We can further coarsen the classification by binning requests with similar lengths. At each time step, the arrival of requests of class $k$ is according to a distribution $\mathbb P_k(\cdot)$ with (finite) expectation $\lambda_k$. Here we impose the following assumption.
\begin{assumption}\label{as1}
    The distribution $\mathbb P_k(\cdot)$ has a finite second moment for any $k\in [m]$.
\end{assumption}

In this framework, we are able to derive the following results.
\begin{proposition}\label{prop1}
    The system cannot be stable when the following inequality holds:
    \begin{equation}\label{eq1}
        \sum_{k=1}^{m}\lambda_k \left(l^{(k)} o^{(k)} + \frac{1}{2} \left(o^{(k)} + (o^{(k)})^2\right) \right) > M.
    \end{equation}
    
\end{proposition}

Proposition \ref{prop1} characterizes a sufficient condition for system instability, or equivalently, a necessary condition that any stable system must satisfy. In Equation \eqref{eq1}, $l^{(k)} o^{(k)} + \frac{1}{2} \left(o^{(k)} + (o^{(k)})^2\right)$ represents the (minimum) overall KV cache consumed by a request of type $k$, measuring the workload. When the inequation holds, the request arrival load exceeds the service capacity, meaning that \emph{no} LLM inference scheduling algorithms can keep the system stable. This proposition bounds the system's ability to complete requests from above.

\begin{algorithm}[t]
\caption{Flow-Controlled LLM Inference Policy with Known Output Length}\label{alg1}
\footnotesize
\begin{algorithmic}
\State \textbf{Input:} KV cache capacity $M$; time horizon $T$; arrival stream $\{\mathcal{A}_t\}_{t=1}^T$.
\State Initialize waiting queue $\mathcal{Q}\leftarrow \emptyset$, active set $\mathcal{P}\leftarrow \emptyset$. \Comment{Initialize}
\For{$t=1,2,\dots,T$}
    \State $\mathcal{Q} \leftarrow \mathcal{Q} \cup \mathcal{A}_t$ \Comment{add new arrivals}
    \State $\mathcal{U}_{t,k} \leftarrow \emptyset$ \Comment{requests upon activated at time $t$}
    \For{$k=1,2,\dots,m$}
        \While{$|\mathcal{U}_{t,k}| < b_k$ \textbf{and} $\mathcal{Q}\neq \emptyset$}
            \State Pick a candidate request $p$ of type $k$ from $\mathcal{Q}$ following the FIFO rule 
            \State $\mathcal{U}_{t,k} \leftarrow \mathcal{U}_{t,k} \cup \{p\}$ \Comment{activate requests when available}
            \State $\mathcal{Q} \leftarrow \mathcal{Q} - \{p\}$
        \EndWhile
        \State $\mathcal{P} \leftarrow \mathcal{P} \cup \mathcal{U}_{t,k}$ \Comment{activate selected requests}
    \EndFor
    \State Process all requests in $\mathcal{P}$
    \State Remove completed requests from $\mathcal{P}$
\EndFor
\end{algorithmic}
\end{algorithm}

In LLM serving, the KV cache usage of each request grows during decoding and is released upon completion. This behavior creates fluctuations in memory utilization. In what follows, we provide a sufficient condition for our flow-controlled LLM inference scheduling policy to achieve stability. 

\begin{theorem}\label{thm1}
    Assume Assumption \ref{as1} holds. Then our flow-controlled LLM inference scheduling policy can achieve stability when the following inequality holds:
    \begin{equation}\label{eq2}
    \sum_{k=1}^{m}b_k \left(l^{(k)} o^{(k)} + \frac{1}{2} \left(o^{(k)} + (o^{(k)})^2\right) \right) < M,
    \end{equation}
where $b_k> \lambda_k$ are integers. Pseudocode is given in Algorithm~\ref{alg1}.
\end{theorem}

The design principle of Algorithm \ref{alg1} is to control the rate at which requests join the active queue. At each time period, the system allows up to $b_k$ requests of type $k$ to join the active set. By doing so, we can make sure that the memory usage at any time remains below the available KV cache capacity. Moreover, since the (maximum) admission rate is larger than the request arrival rate, we can show that the system can be stable. The proof is provided in Appendix \ref{app1}.

\subsection{Unknown Output Length}\label{sec3.3}
In this section, we consider a more realistic setting, where the output length is unknown upon request arrival. For any request $p_i$, we now assume that the input and output lengths $(l_i,o_i)$ are stochastic and are i.i.d. draws from a distribution $\nu$. Upon arrival, $l_i$ is observed while $o_i$ is unknown. The system needs to make decisions based on the knowledge of $l_i$ and the distribution of $o_i$. To facilitate our analysis, we impose the following assumption, which assumes that all input and output lengths are bounded:
\begin{assumption}\label{as2}
    There exists a universal constant $C>0$ such that $l,o\leq C$ holds for any $(l,o)$ drawn from the distribution $\nu$.
\end{assumption}
This assumption is mild in practice because the input and output lengths are typically bounded by the architectural limits of the model. 

We first establish the following results that provide a sufficient condition for instability.

\begin{proposition}\label{prop2}
    Assume Assumption \ref{as2} holds. Then the system cannot be stable when the following inequality holds:
    \begin{equation}\label{eq3}
        \lambda \mathbb E\left[l o + \frac{1}{2} \left(o + o^2\right) \right] > M,
    \end{equation}
    where $\lambda$ is the request arrival rate, and $(l,o)\sim \nu$.
\end{proposition}

Proposition \ref{prop2} implies that the queue of waiting requests will explode if the average request arrival rate exceeds a threshold. In what follows, we formally propose our policy Algorithm \ref{alg2} for the scenario where the output length is unknown. This algorithm builds upon Algorithm \ref{alg1} by incorporating an additional (heuristic) eviction rule to handle the overflow event. In our paper, when overflow occurs, we choose to evict active requests following the last-in, first-out (LIFO) mechanism until the memory usage does not exceed the limit. Intuitively, LIFO eviction prioritizes keeping requests that have been decoding the longest, which have already consumed significant KV cache resources during their decoding. Thus, evicting them would waste substantial computational effort. Moreover, requests that are close to their completion will soon release their entire memory footprint. By contrast, recently activated requests have generated few tokens. Evicting them can incur minimal wasted work. This makes LIFO a natural choice for balancing overflow recovery with resource efficiency. We then present the following theoretical guarantees for this policy, which demonstrates that the probability of memory overflow decays with the slack between the expected memory usage and the limit.

\begin{algorithm}[t]
\caption{Flow-Controlled LLM Inference Policy with Unknown Output Length}\label{alg2}
\footnotesize
\begin{algorithmic}
\State \textbf{Input:} KV cache capacity $M$; time horizon $T$; activation budget $\{B_t\}_{t=1}^T$; arrival stream $\{\mathcal{A}_t\}_{t=1}^T$.
\State Initialize waiting queue $\mathcal{Q}\leftarrow \emptyset$, active set $\mathcal{P}\leftarrow \emptyset$. \Comment{Initialize}
\For{$t=1,2,\dots,T$}
    \State $\mathcal{Q} \leftarrow \mathcal{Q} \cup \mathcal{A}_t$ \Comment{add new arrivals}
    \State $\mathcal{U}_{t} \leftarrow \emptyset$ \Comment{requests upon activated at time $t$}
    \While{$|\mathcal{U}_{t}| < B_t$ \textbf{and} $\mathcal{Q}\neq \emptyset$}
        \State Pick a candidate request from $\mathcal{Q}$ following the FIFO rule
        \State $\mathcal{U}_{t} \leftarrow \mathcal{U}_{t} \cup \{p\}$ \Comment{activate requests when available} 
        \State $\mathcal{Q} \leftarrow \mathcal{Q} - \{p\}$
    \EndWhile
    \State $\mathcal{P} \leftarrow \mathcal{P} \cup \mathcal{U}_t$ \Comment{activate selected requests (enter active set)}
    \If{\textsc{Overflow}}
        \State $E\leftarrow  \textsc{Evict}(\mathcal{P})$ \Comment{select which requests to evict}
        \State $\mathcal{P}\leftarrow \mathcal{P} - E$ \Comment{eviction when the memory usage exceeds the budget}
        \State $\mathcal{Q} \leftarrow \mathcal{Q} \cup E$
    \EndIf
    \State Process all requests in $\mathcal{P}$
    \State Remove completed requests from $\mathcal{P}$
\EndFor
\end{algorithmic}
\end{algorithm}

\begin{theorem}\label{thm2}
    Assume Assumption \ref{as2} holds, and there exists $\epsilon>0$, such that
    \begin{equation}\label{eq}
    b \mathbb E\left[l o + \frac{1}{2} \left(o + o^2\right) \right] \leq (1-\epsilon) M
    \end{equation}
    holds for $b> \lambda$. Let $\{B_t\}_{t=1}^T$ be i.i.d. nonnegative integer-valued random variables that are independent of the requests and satisfy
    \begin{align*}
        \mathbb E [B_t] = b,\quad B_t \leq A
    \end{align*}
    holds for any $t=1,\ldots,T,$ $A>0$. Then under our proposed Algorithm~\ref{alg2}, the expected number of overflow events is less than $T\exp(-C(A,\epsilon)M^2)$ for some $C(A,\epsilon)>0$.
\end{theorem}
In practice, the performance of our algorithm can be sensitive to the choice of $b$. When $b$ is small, the algorithm is conservative and may limit the memory usage. When $b$ is large, the algorithm gradually degenerates to a greedy algorithm with a LIFO eviction rule. We can perform a grid search to find a proper $b$, allowing for $b$ that does not satisfy Inequality \eqref{eq}.

\section{Numerical Simulations}\label{sec4}
In this section, we evaluate our proposed flow-controlled algorithm through numerical simulations. We measure its performance using the two performance metrics we discussed in Section \ref{sec2.2}: throughput, and latency. We consider two simulation settings. In Section \ref{sec4.1}, we conduct simulations on synthetic data and compare the performance of our algorithm against some baseline heuristic scheduling algorithms. In Section \ref{sec4.2}, we conduct simulations on a large-scale real-world dataset.

\subsection{Synthetic Data Simulations}\label{sec4.1}
We first evaluate the performance of our algorithm on synthetic datasets. We consider a simulation setting where the simulated prompts can be classified into $m=3$ types according to the input and output lengths: 
\[(l_1,o_1) = (10,20),\quad (l_2,o_2) = (10,40),\quad (l_3,o_3) = (10,60).\]
These classes share the same input length but differ in output length. Each type of request arrives as an independent Poisson process with the same intensity $\lambda$. We allow the time horizon $T$ to vary with $T \in \{1000,2000,\ldots,10000\}$, and let $\lambda = 5$. We simulate a single-GPU inference system with a hard KV cache capacity $M$. We set $M=16492$, which matches the deployment scale for Llama2-70B on two A100 GPUs. In this scenario, we set $(b_1,b_2,b_3)=(4,4,4)$ so that Inequality \eqref{eq2} is met. This choice makes sure that the memory usage will never exceed the limit.

\begin{figure}
  \centering
  \begin{minipage}{1\textwidth}
    \centering
    \begin{minipage}[t]{0.24\linewidth}
      \centering
      \includegraphics[width=\linewidth]{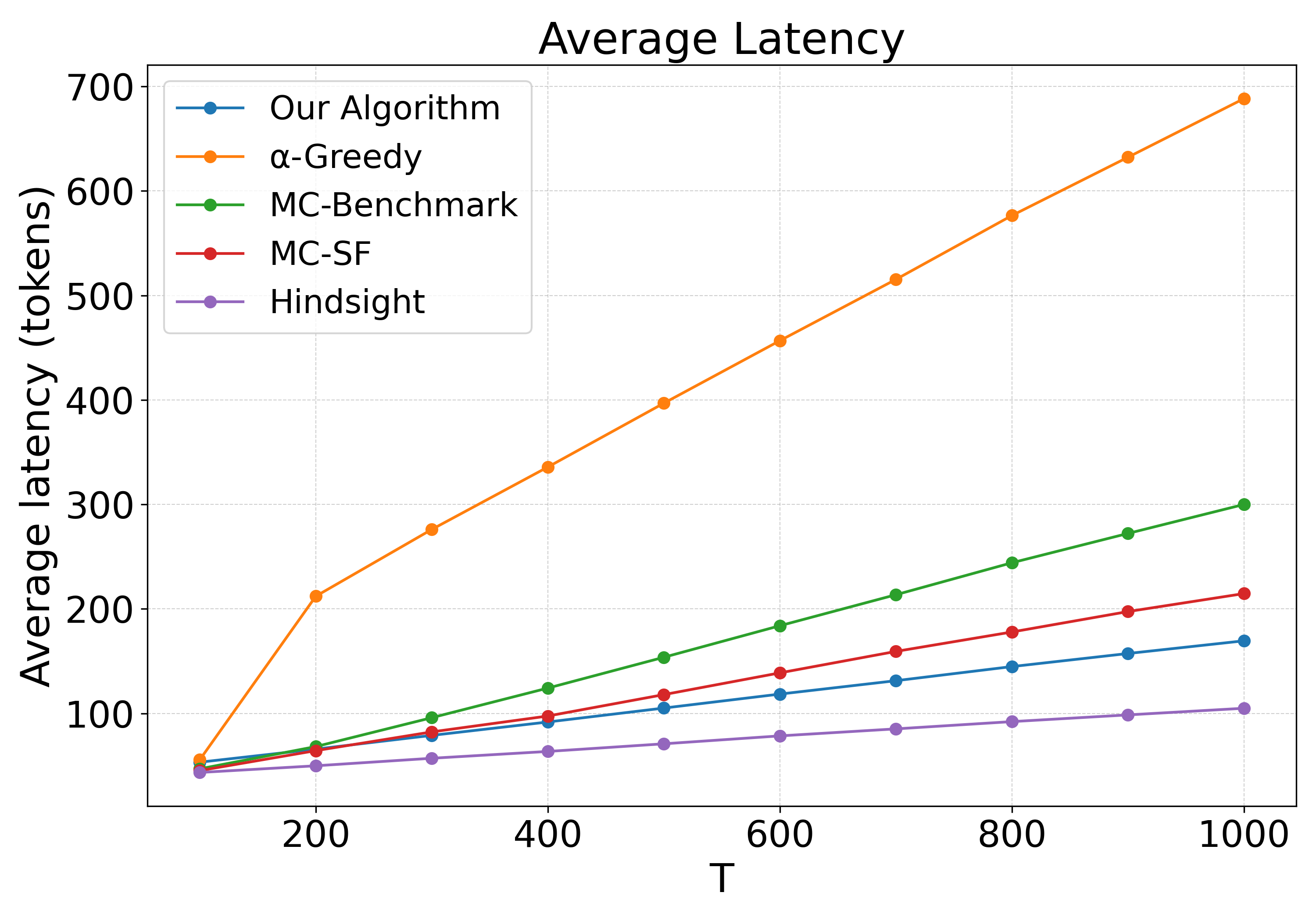} 
    \end{minipage}\hfill
    \begin{minipage}[t]{0.24\linewidth}
      \centering
      \includegraphics[width=\linewidth]{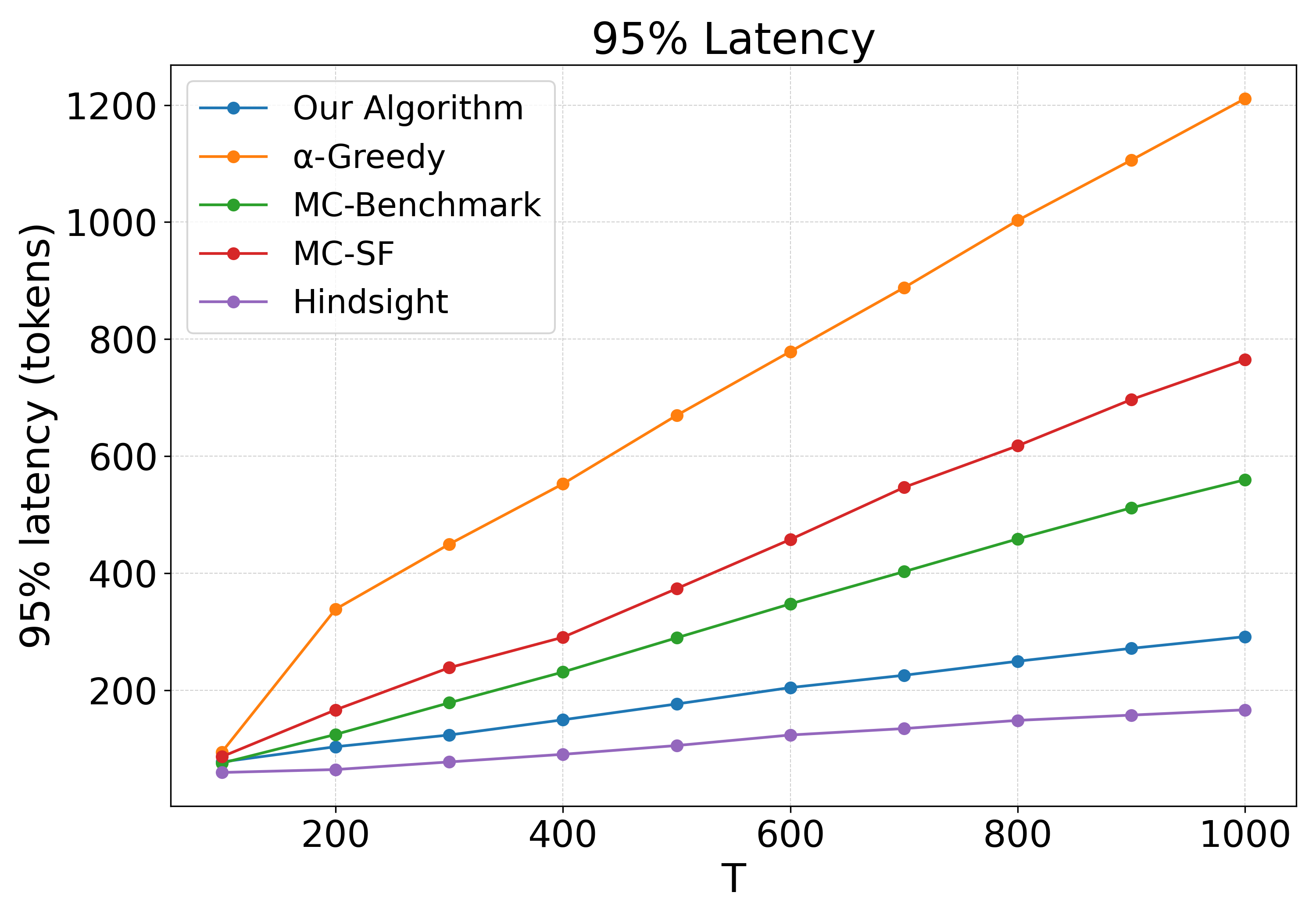} 
    \end{minipage}\hfill
    \begin{minipage}[t]{0.24\linewidth}
      \centering
      \includegraphics[width=\linewidth]{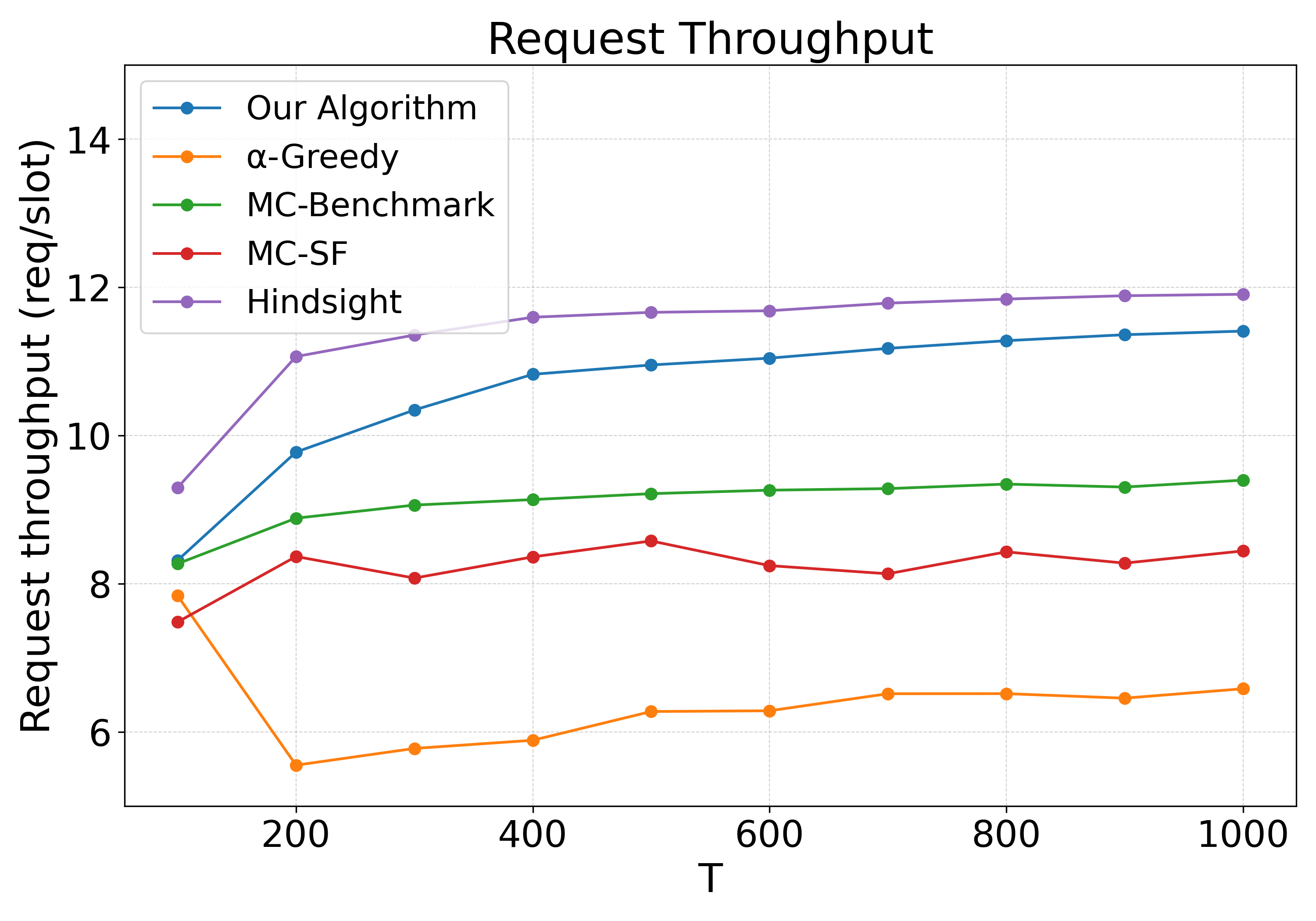} 
    \end{minipage}
    \begin{minipage}[t]{0.24\linewidth}
      \centering
      \includegraphics[width=\linewidth]{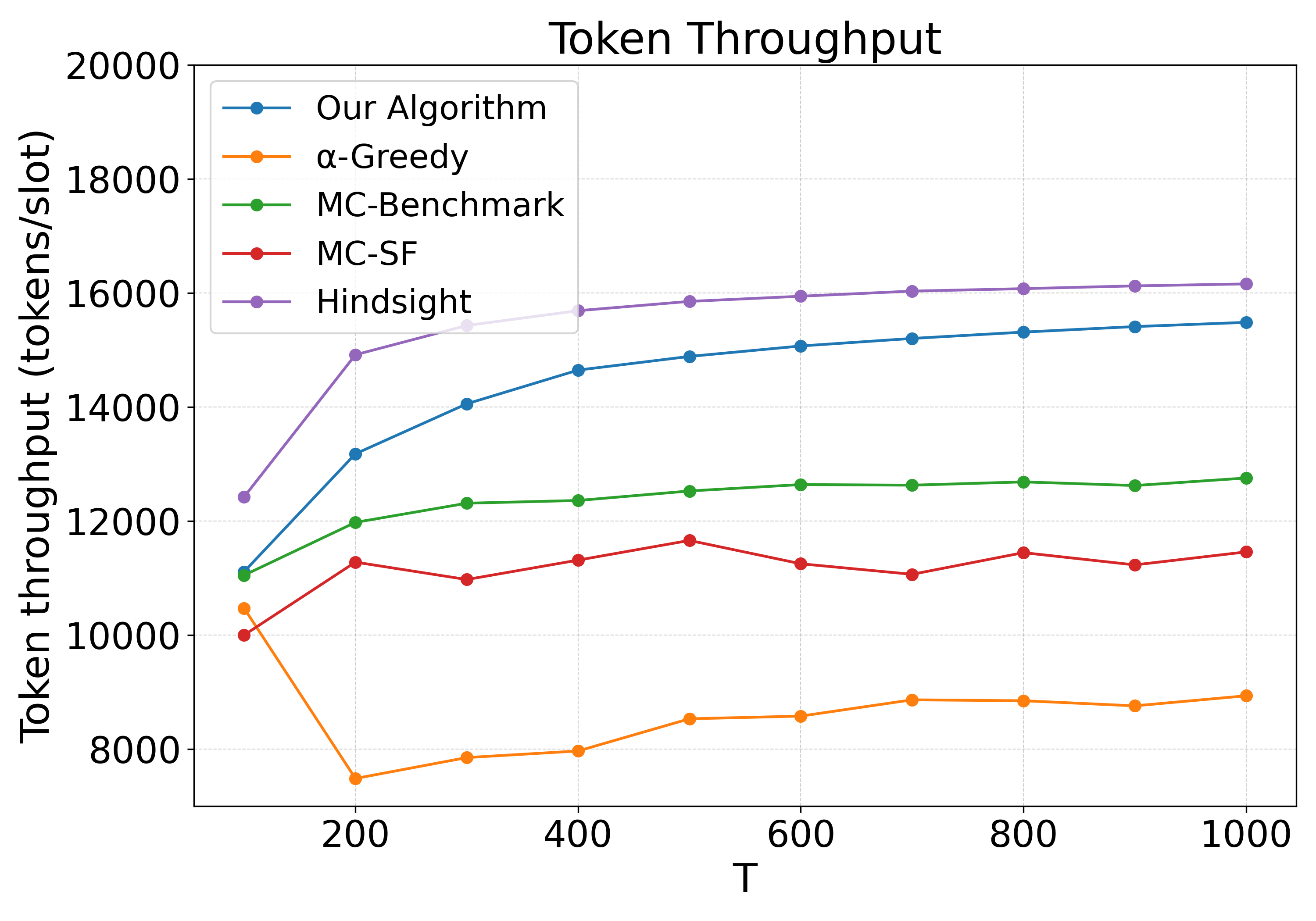} 
    \end{minipage}
  \end{minipage}

  \begin{minipage}{1\textwidth}
    \centering
    \begin{minipage}[t]{0.248\linewidth}
      \centering
      \includegraphics[width=\linewidth]{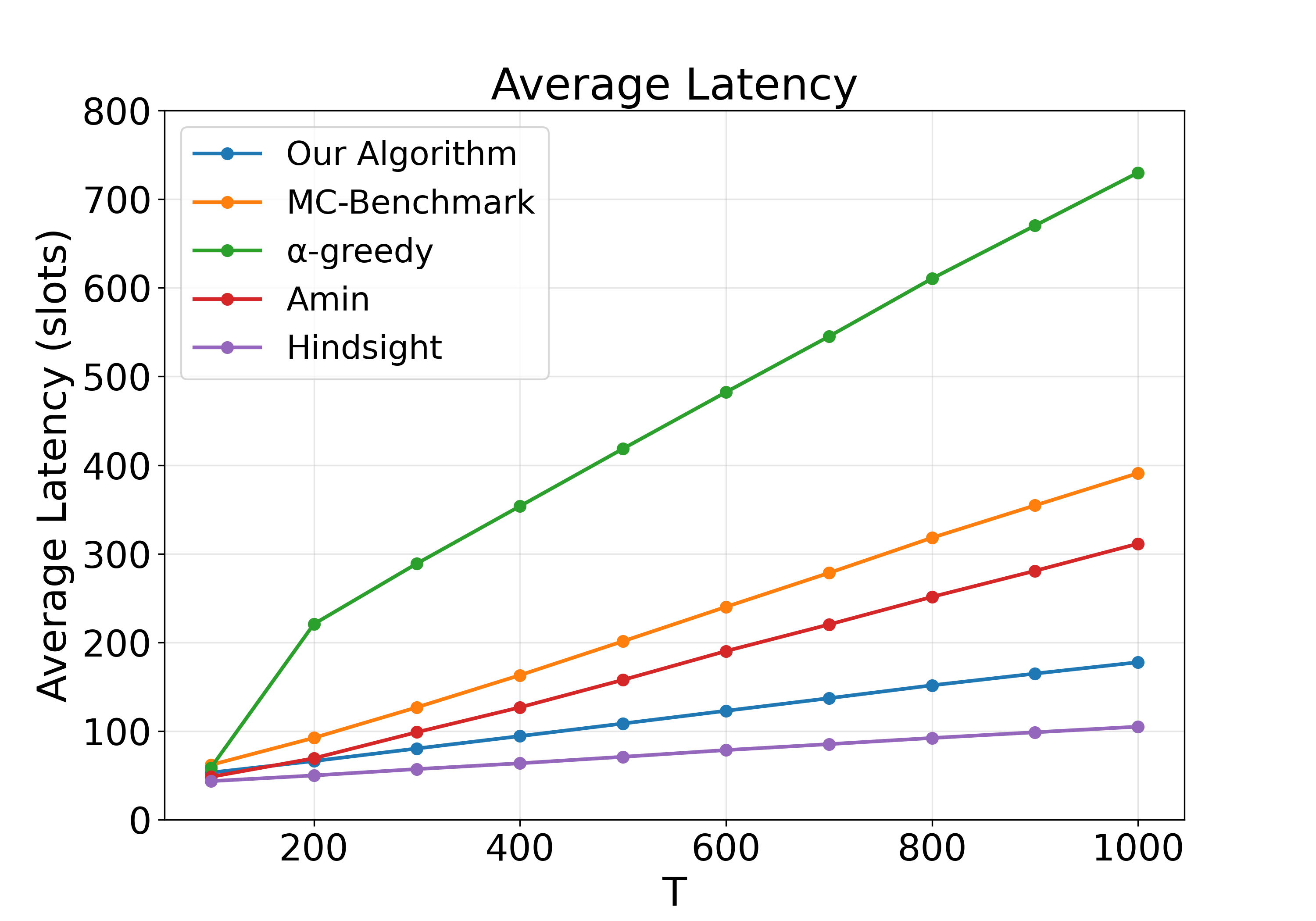} 
    \end{minipage}\hfill
    \begin{minipage}[t]{0.248\linewidth}
      \centering
      \includegraphics[width=\linewidth]{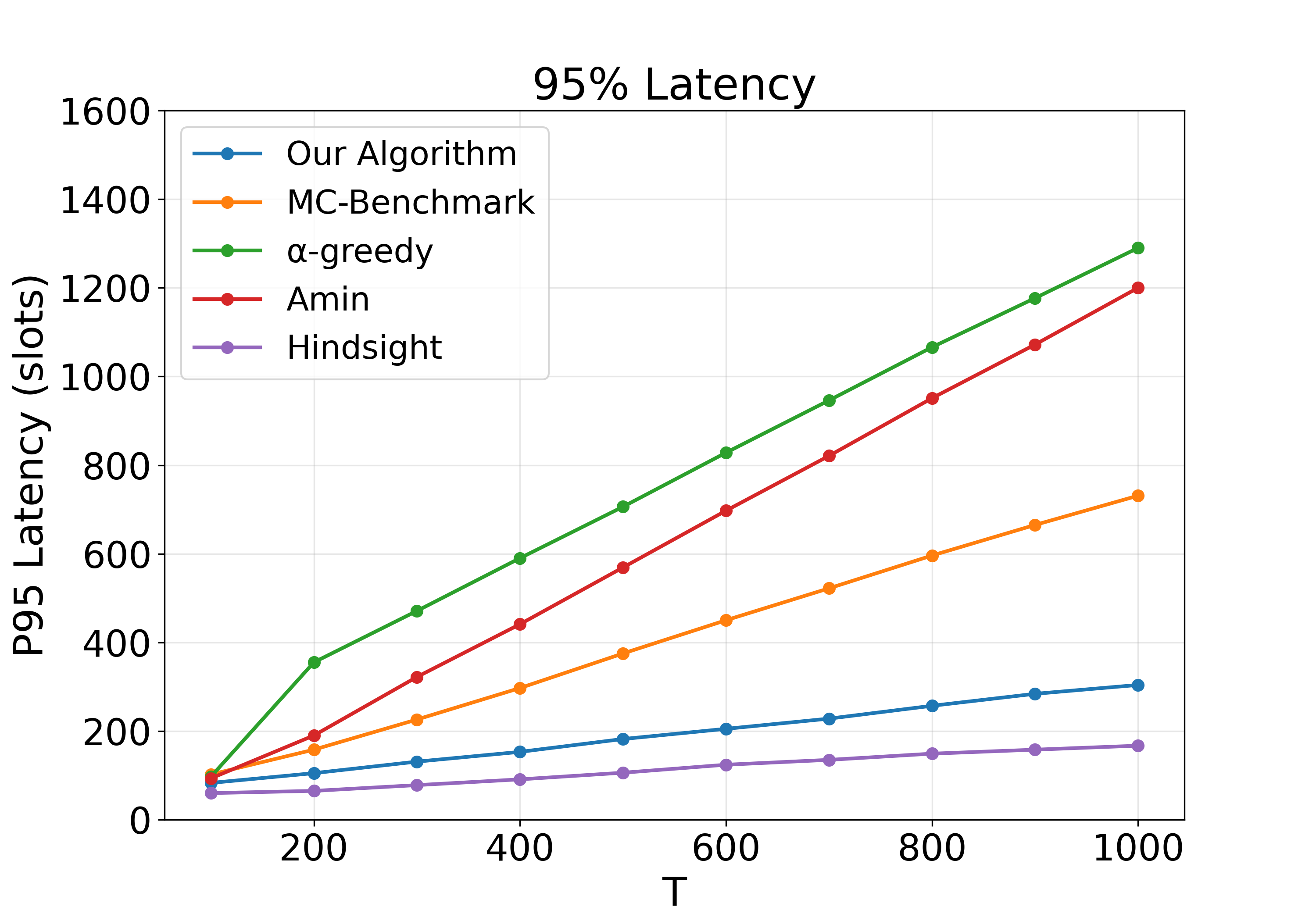} 
    \end{minipage}\hfill
    \begin{minipage}[t]{0.248\linewidth}
      \centering
      \includegraphics[width=\linewidth]{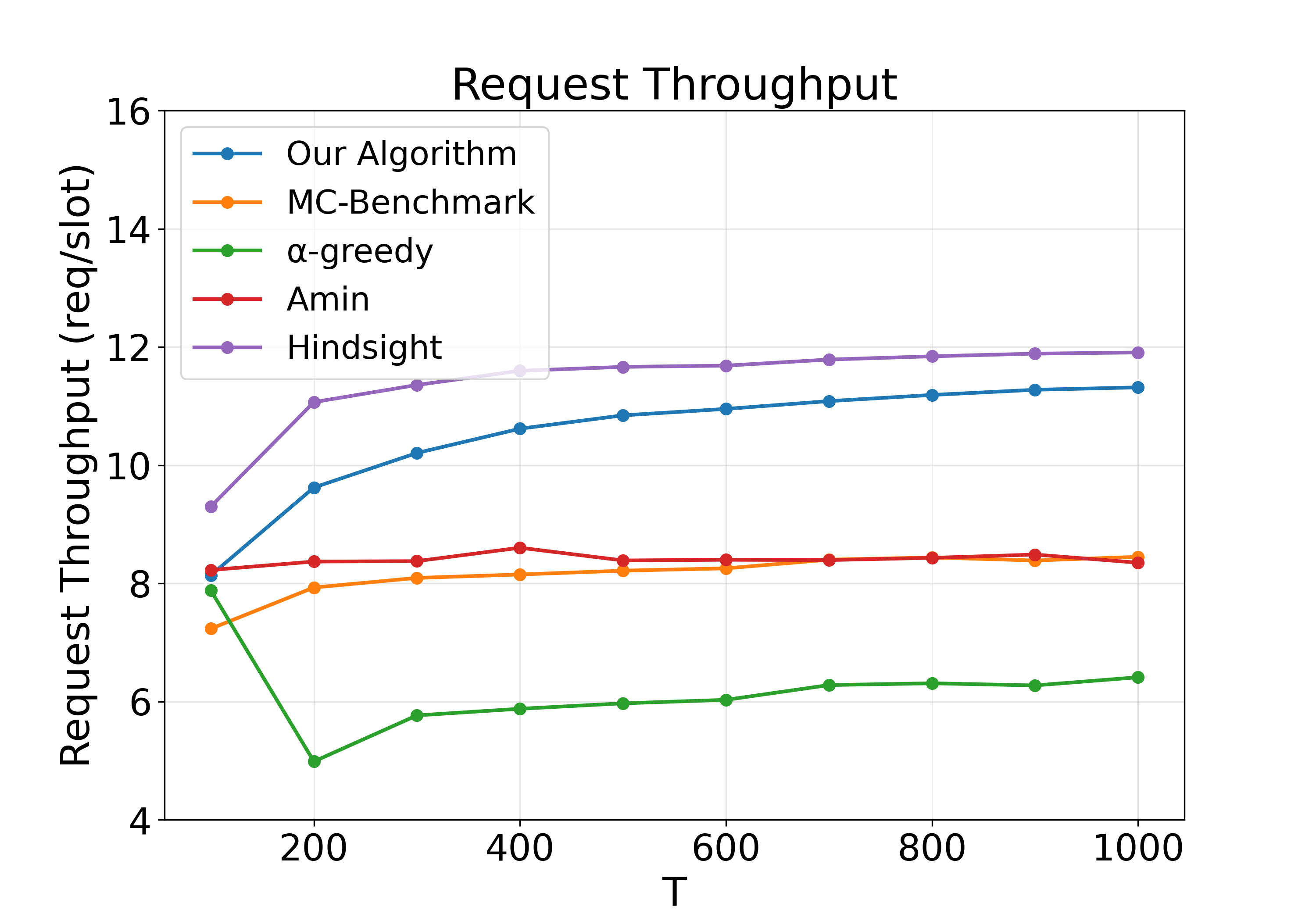} 
    \end{minipage}
    \begin{minipage}[t]{0.248\linewidth}
      \centering
      \includegraphics[width=\linewidth]{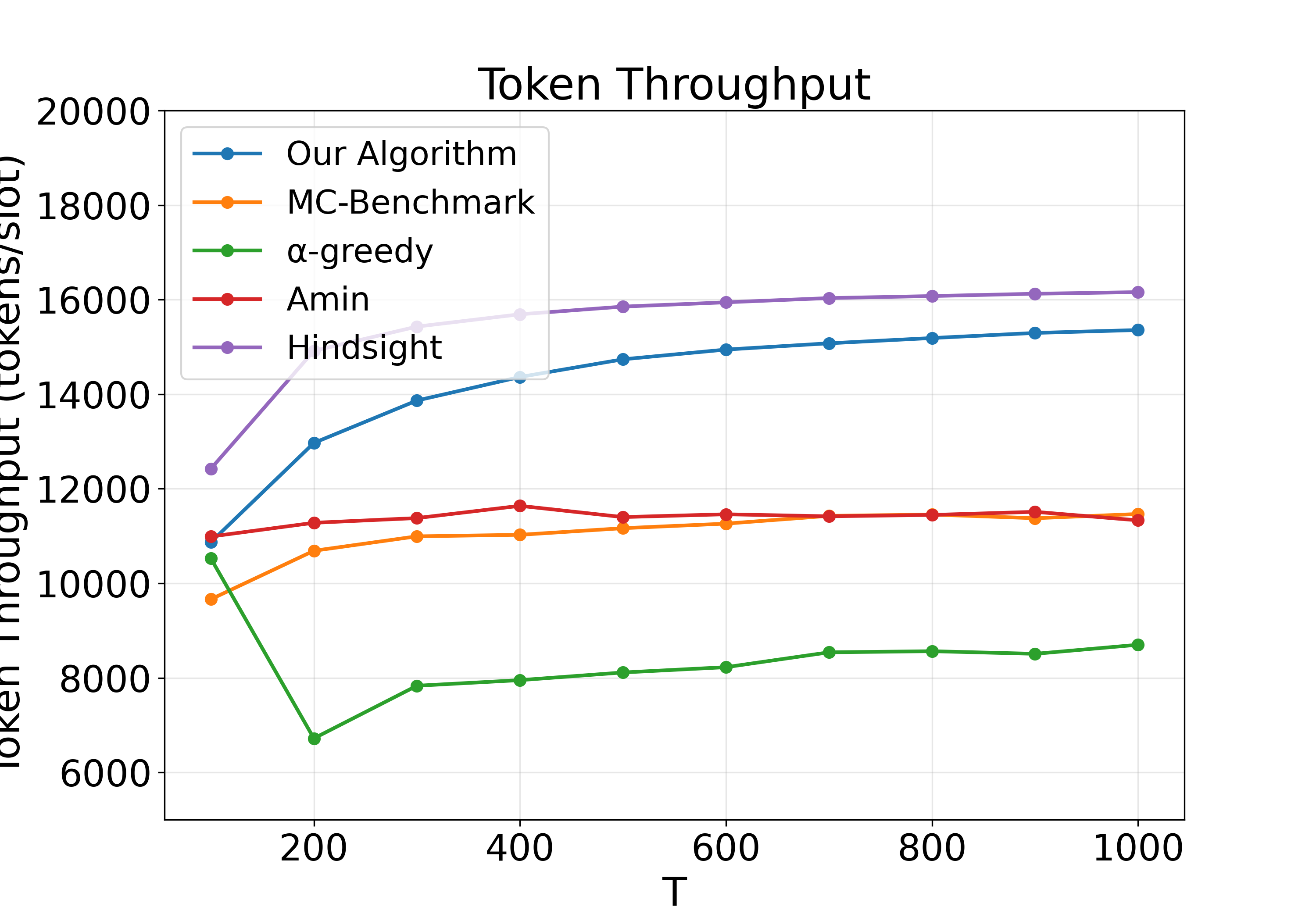} 
    \end{minipage}
  \end{minipage}

\caption{(Up) Performance metrics across scheduling algorithms when the request type is known upon arrival. (Down) Performance metrics across scheduling algorithms when the request type is unknown upon arrival. The performance metrics (from left to right) are average latency, 95\% latency, request throughput, and token throughput.}\label{f1} 
\end{figure}

\subsubsection{Results}
To begin with, we assume that the request type is known upon request arrival. We compare our algorithm against a set of baselines that capture widely used scheduling heuristics under KV cache constraints and compute the performance metrics, i.e., average end-to-end latency, 95\% latency, request throughput, and token throughput.
\begin{itemize}
    \item \emph{$\alpha$-protection greedy algorithms (\citep{jaillet2025online}).} This class of algorithms maintains a protection memory threshold of $\alpha M$. The algorithm prioritizes existing token jobs. For a new prompt $p_i$, if adding it with its initial memory $l_i+1$ would not cause the memory usage to exceed the allowable limit $(1-\alpha)M$, then the algorithm will add it in the active set. Otherwise, this prompt will not be added. The algorithm evicts all active requests and sends them back to the waiting queue upon overflow. Here, we perform a grid search on the $\alpha$-protection greedy heuristic and take a proper value of $\alpha$.
    \item \emph{MC benchmark.} This is an FCFS-style heuristic algorithm, in which the algorithm only activates requests if the future memory usage remains within the memory limit.
    \item \emph{MC-SF.} This algorithm is proposed by \cite{jaillet2025online}. This algorithm prioritizes adding requests with smaller decoding lengths to allow more requests to be packed into batches. The algorithm activates requests when the future memory usage is within the limit.
\end{itemize}
Moreover, to obtain the hindsight-optimal value for each performance metric, we formulate a separate integer programming model and solve it using Gurobi. Since each integer programming model optimizes a different objective, the resulting optimal solutions differ across metrics (i.e., no single schedule is simultaneously optimal for all metrics). The results are presented in the first line of Figure \ref{f1}. Notably, our method outperforms the three benchmarks across all metrics.

Next, we consider the setting in which the output length of each request is unknown upon arrival. In this setting, the MC-SF algorithm becomes less applicable because we do not have access to the decoding length of each request. To this end, we benchmark against the Amin algorithm proposed by \cite{chen2025adaptively}. Concretely, the Amin algorithm treats the decoding length of each request as the shortest possible value and greedily activates requests. Upon overflow, the algorithm removes requests according to the ascending order of the output length predictions and updates the predictions for the evicted requests. By contrast, our strategy is more conservative. Rather than aggressively filling the batch, our method proactively limits the number of activated requests for stability and robustness under output-length uncertainty. Moreover, we also adapt the MC benchmark algorithm to this setting by treating the output lengths of requests as the longest possible value. The results are reported in the second row of Figure \ref{f1}. We observe that our method again outperforms the benchmarks across all metrics and achieves performance close to the hindsight optimal. This demonstrates the strength of our algorithm because it achieves good performance whether the request output length is known or unknown.

\subsection{Real-World Data Simulations}\label{sec4.2}
\begin{figure}
  \centering

  \begin{minipage}{1\textwidth}
    \centering
    \begin{minipage}[t]{0.49\linewidth}
      \centering
      \includegraphics[width=\linewidth]{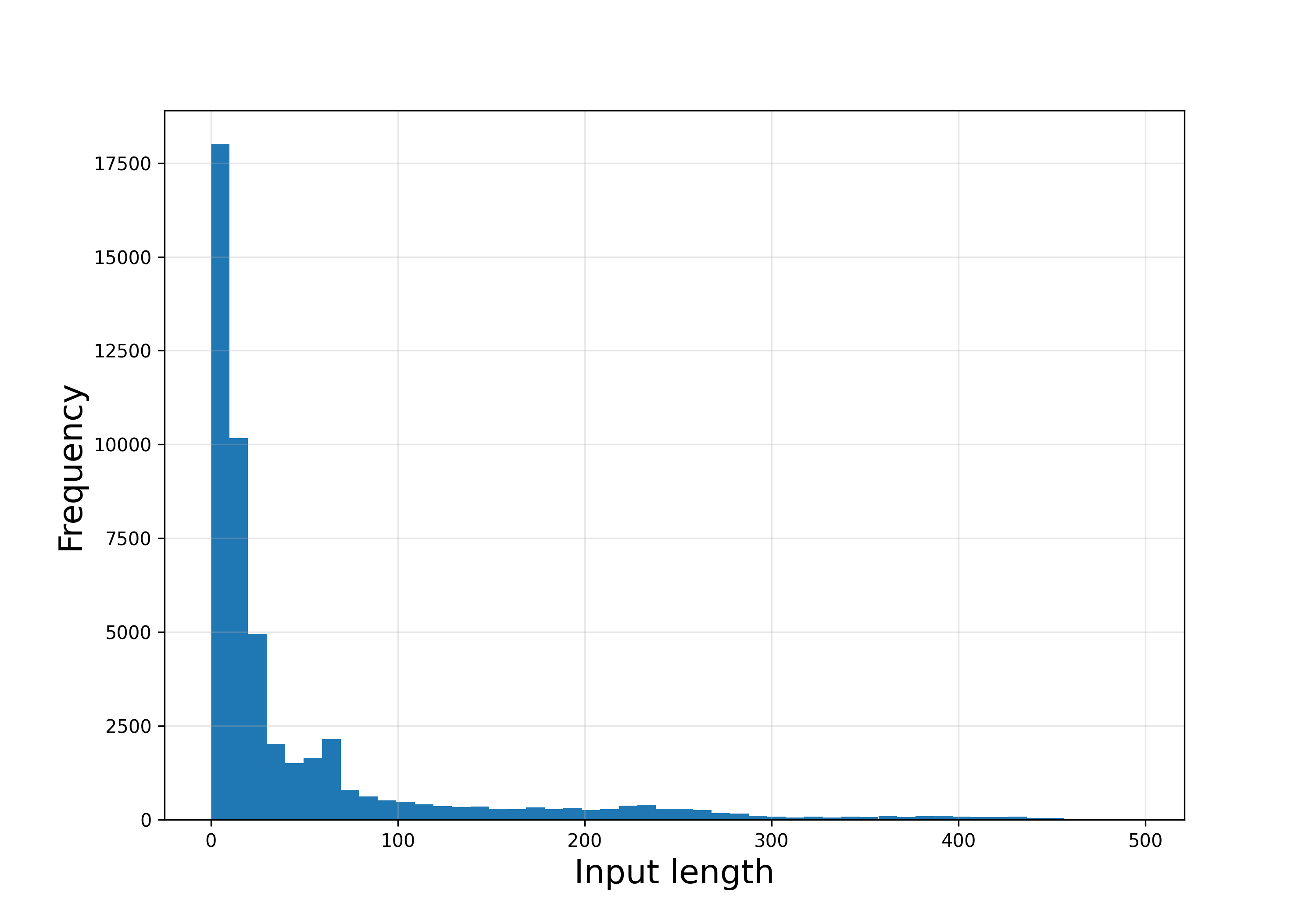} 
    \end{minipage}\hfill
    \begin{minipage}[t]{0.49\linewidth}
      \centering
      \includegraphics[width=\linewidth]{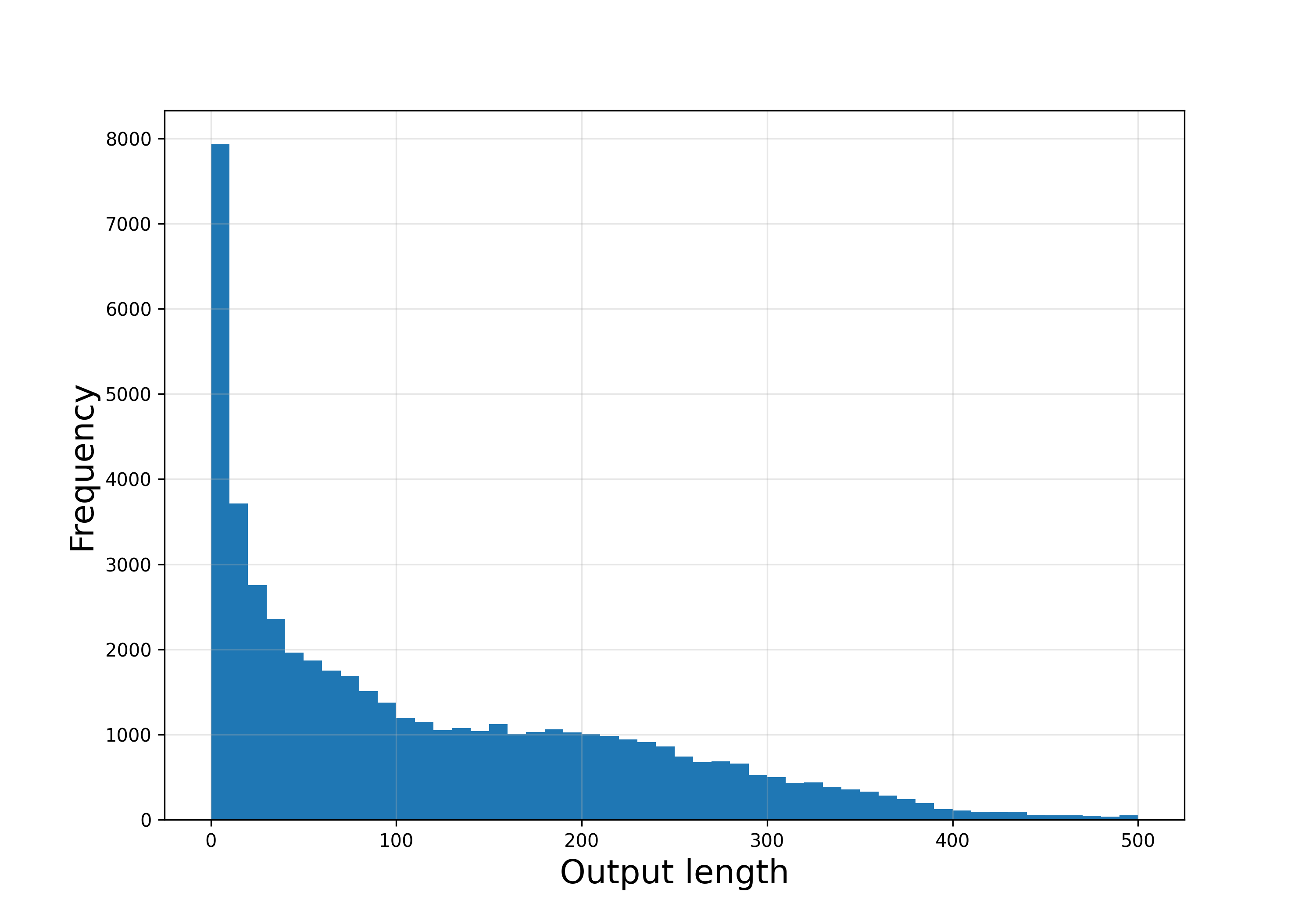} 
    \end{minipage}
  \end{minipage}

\caption{Distribution of prefill and decode lengths in the real-world dataset.}\label{f0}
\end{figure}

In this section, we evaluate the performance of our algorithm on a real-world dataset from \cite{zheng2023lmsys}, which contains over 210000 prompts. We randomly select 50000 prompts from this dataset and treat each as an inference request: the prompt serves as an input, and the response serves as an output. We use word count as a proxy for token length, where each word is defined as a token. The requests are simulated to arrive according to a Poisson process with intensity $\lambda$. We allow $\lambda$ to vary with $\lambda \in \{10,50\}$ (requests per second) to simulate both high- and low-demand settings. Moreover, in our previous model, we discretize the process time of a batch into a time period. However, in practical LLM serving, it depends on the batch composition (e.g., the number of prefill and decode tokens). Therefore, in our numerical results, we simulate a continuous-time serving system and adopt Microsoft Vidur \citep{agrawal2024vidur} to calculate the process time. Figure \ref{f0} displays the distribution of input and output lengths of the requests.

\subsubsection{Results}
\begin{figure}
  \centering

  \begin{minipage}{1\textwidth}
    \centering
    \begin{minipage}[t]{0.24\linewidth}
      \centering
      \includegraphics[width=\linewidth]{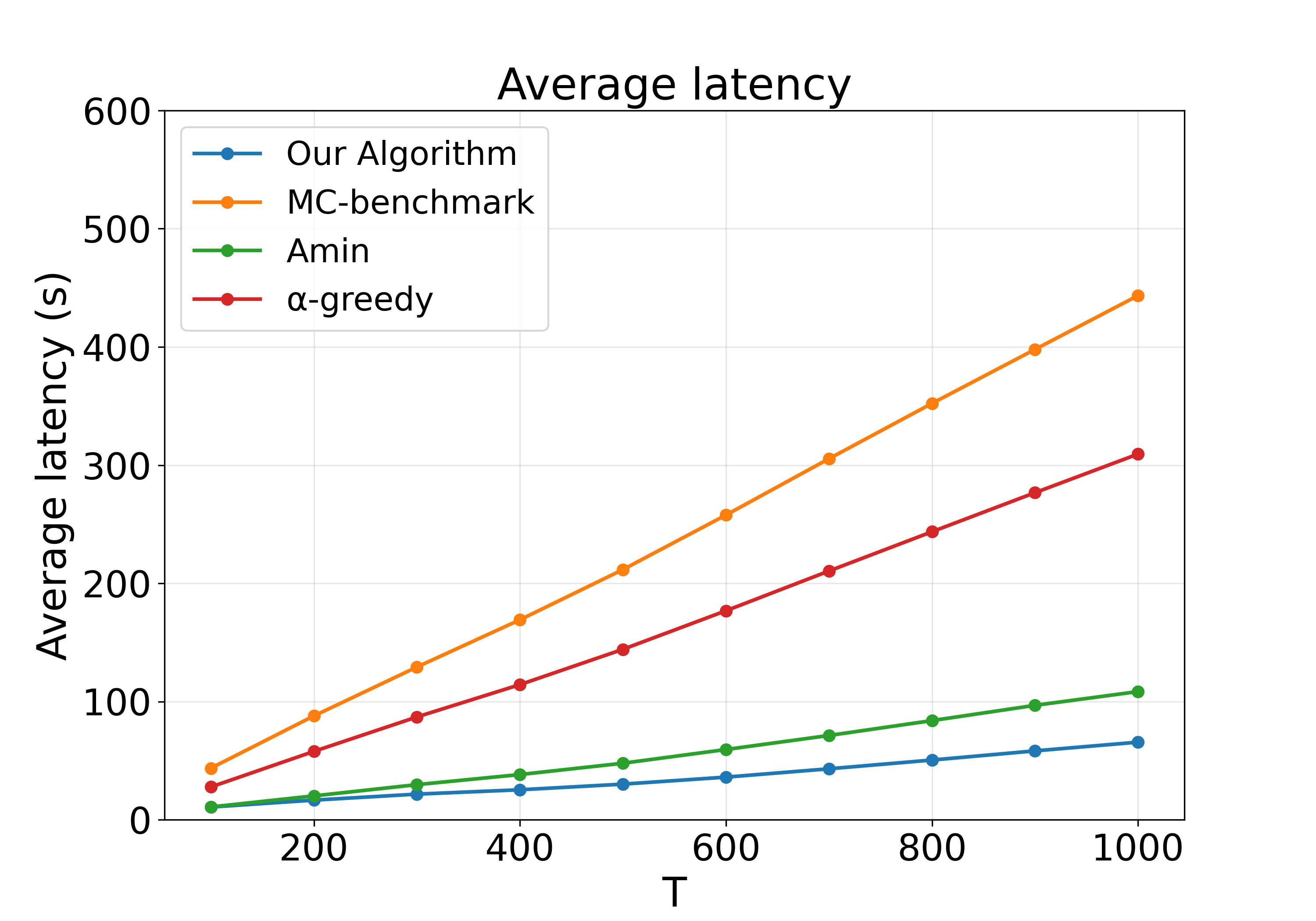} 
    \end{minipage}\hfill
    \begin{minipage}[t]{0.24\linewidth}
      \centering
      \includegraphics[width=\linewidth]{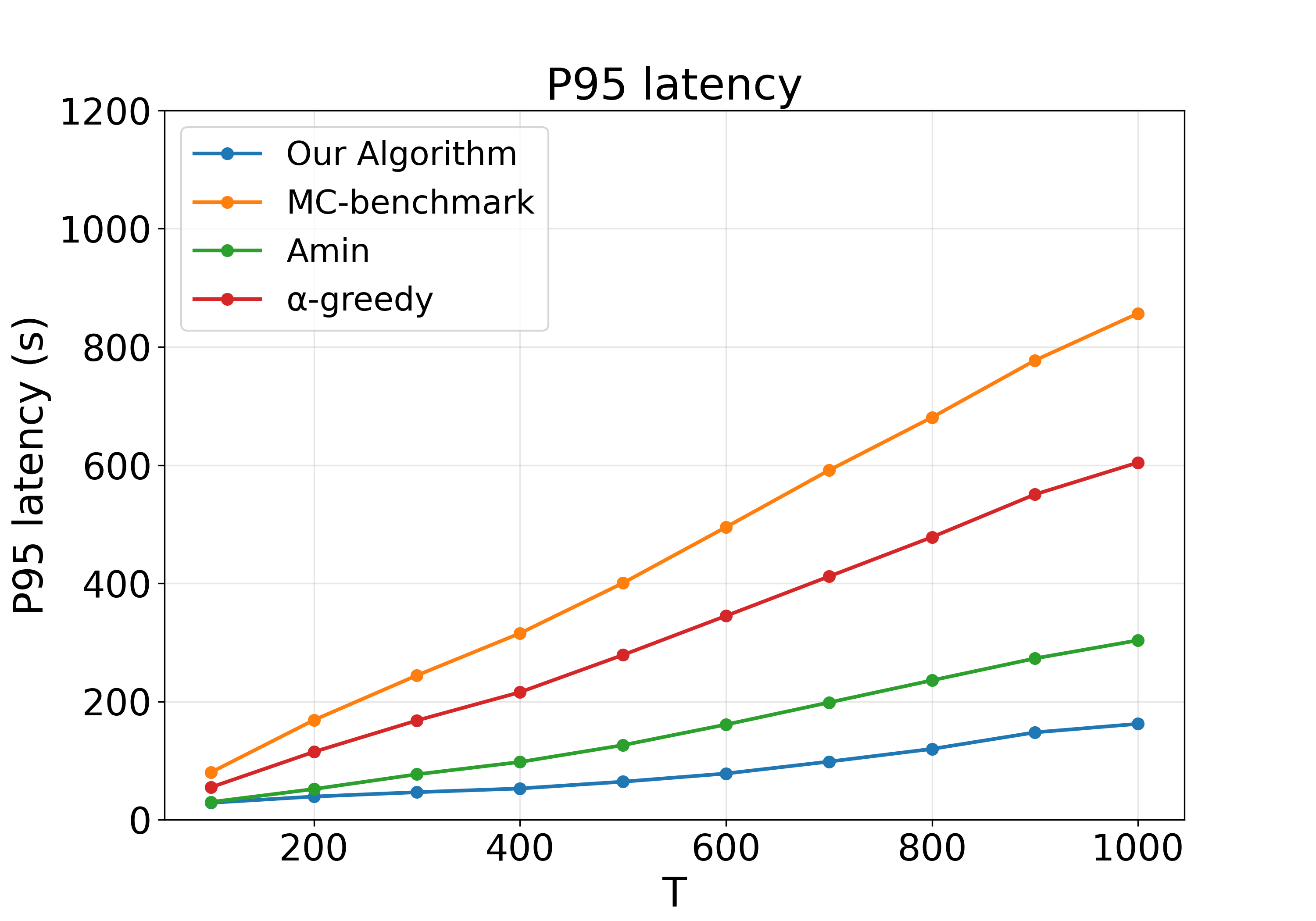} 
    \end{minipage}\hfill
    \begin{minipage}[t]{0.24\linewidth}
      \centering
      \includegraphics[width=\linewidth]{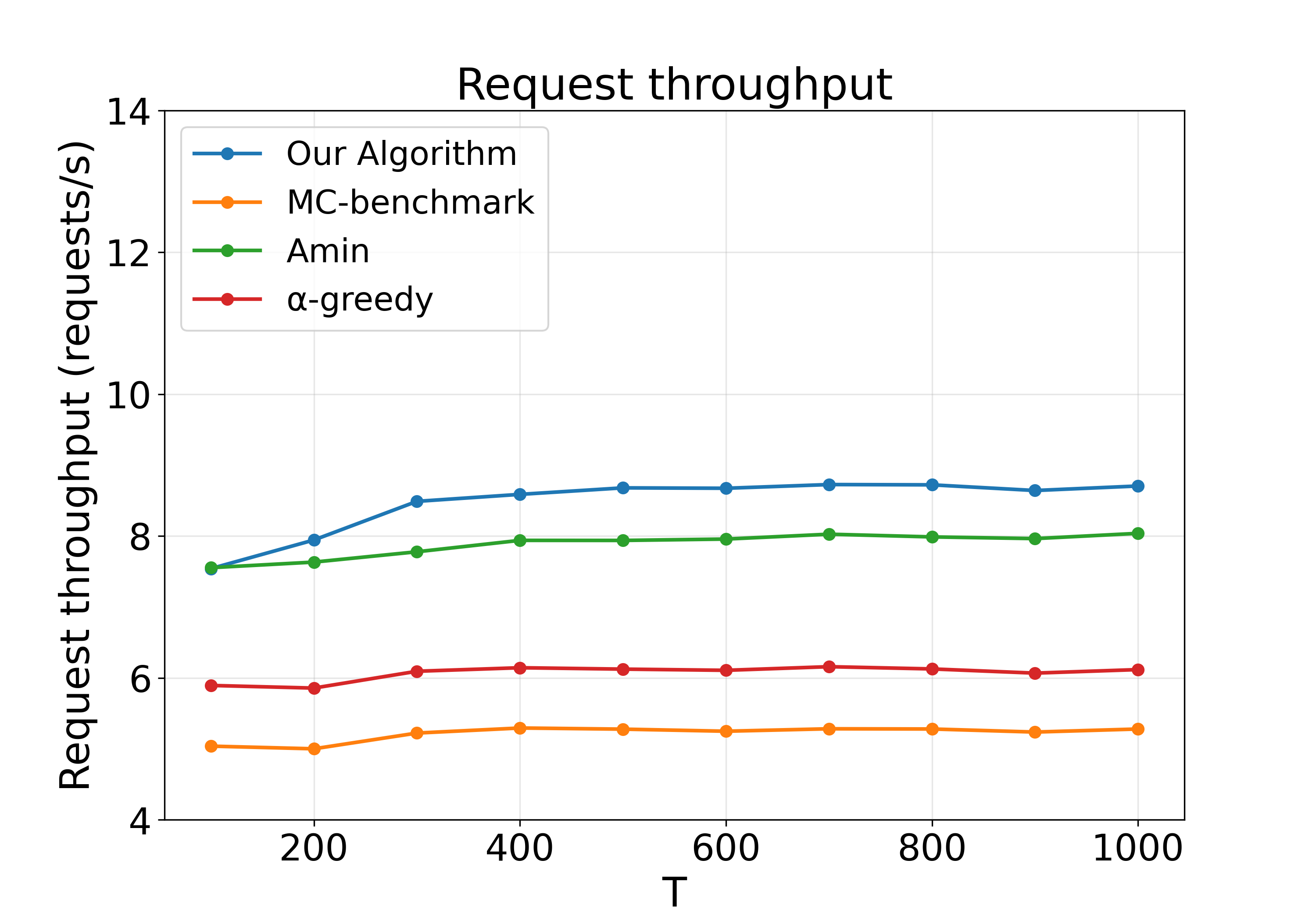} 
    \end{minipage}
    \begin{minipage}[t]{0.24\linewidth}
      \centering
      \includegraphics[width=\linewidth]{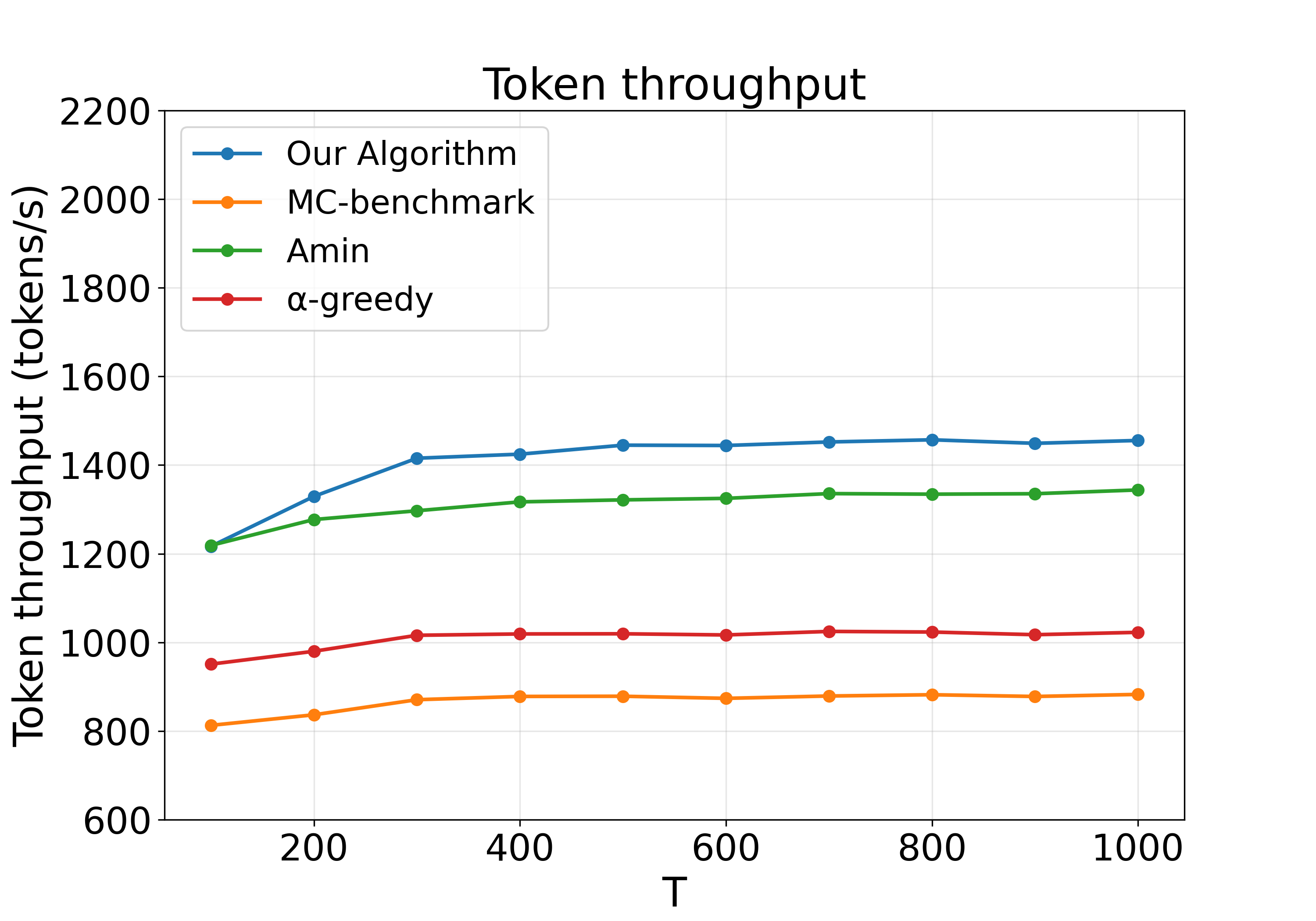} 
    \end{minipage}
  \end{minipage}

  \begin{minipage}{1\textwidth}
    \centering
    \begin{minipage}[t]{0.24\linewidth}
      \centering
      \includegraphics[width=\linewidth]{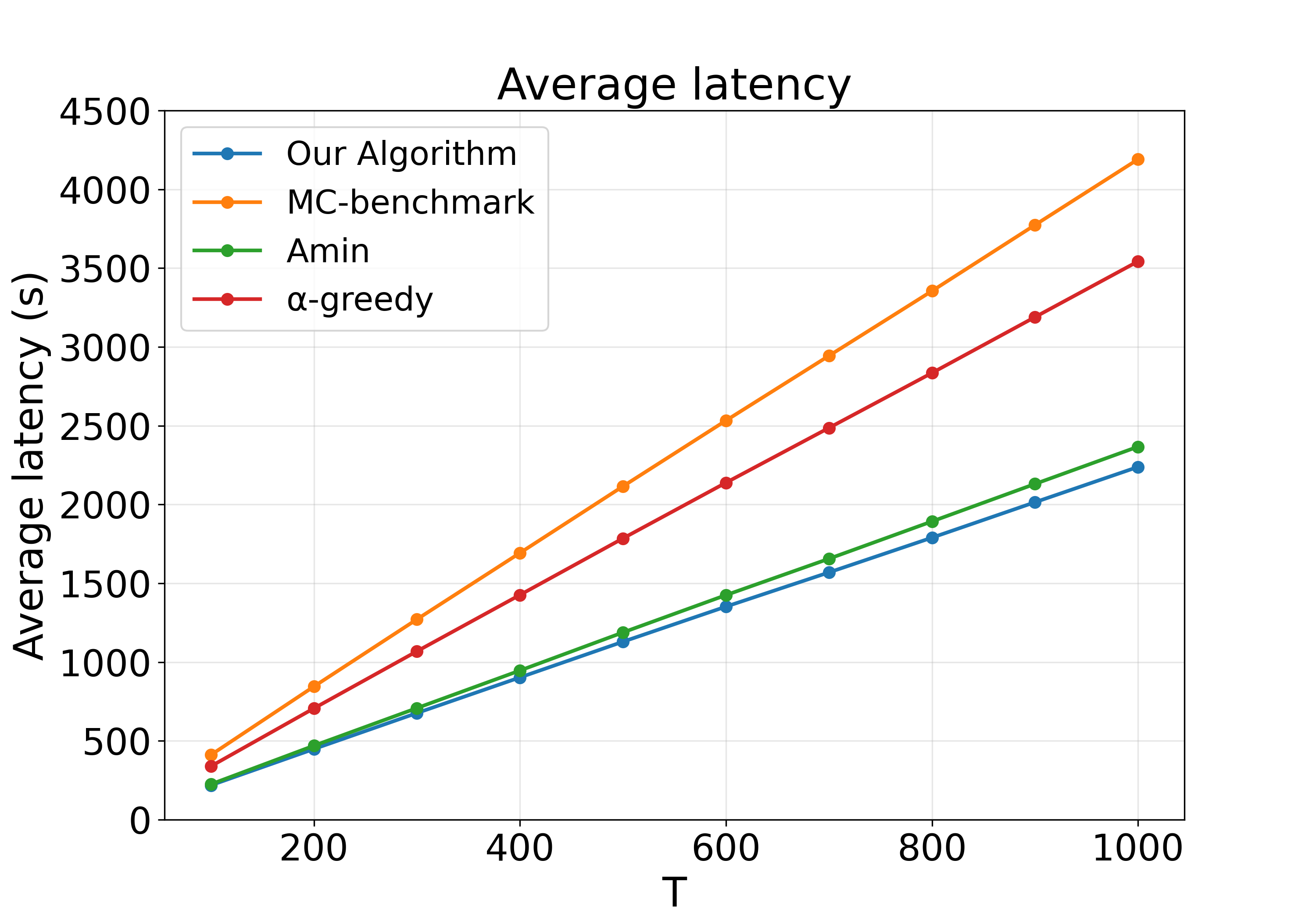} 
    \end{minipage}\hfill
    \begin{minipage}[t]{0.24\linewidth}
      \centering
      \includegraphics[width=\linewidth]{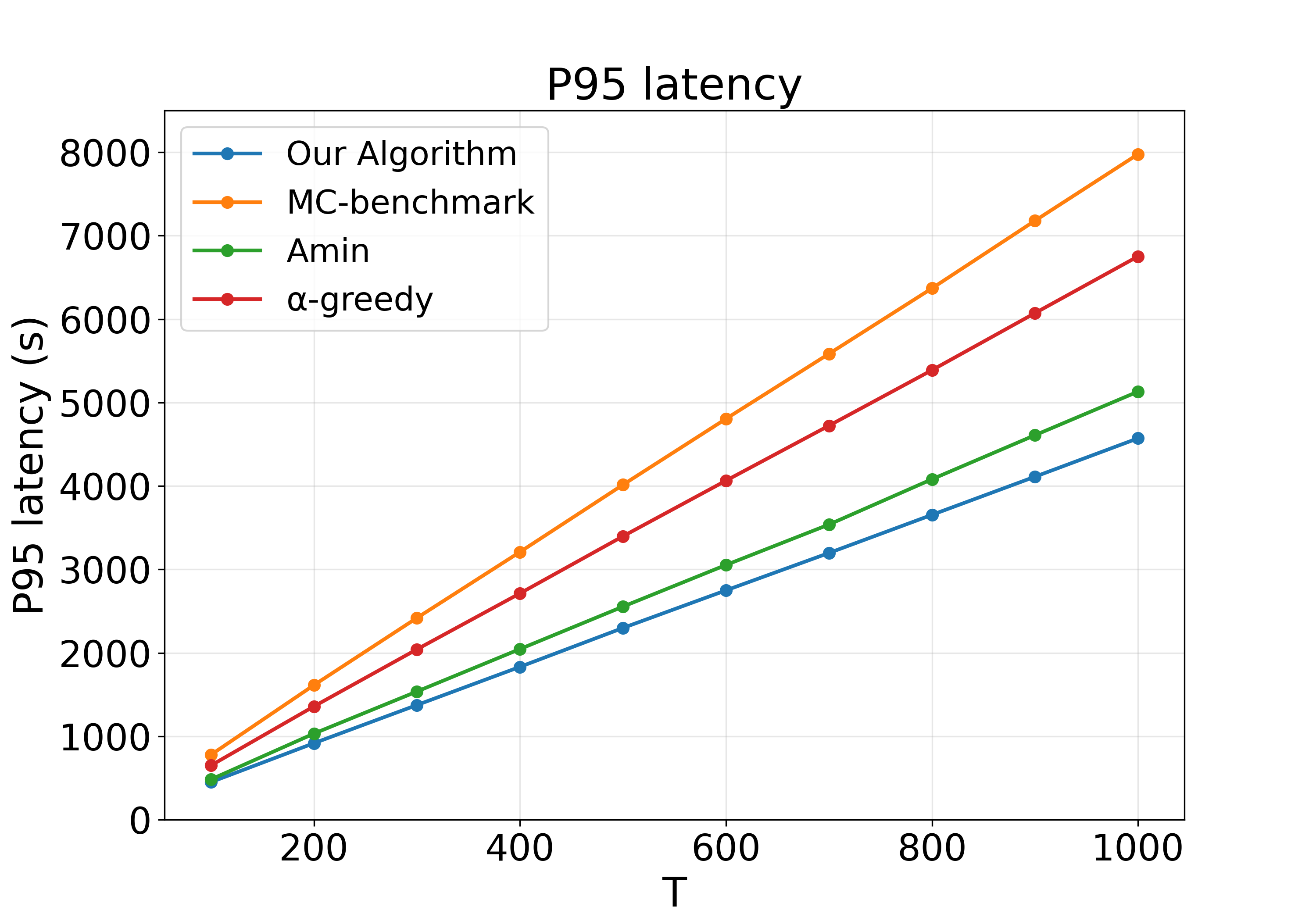} 
    \end{minipage}\hfill
    \begin{minipage}[t]{0.24\linewidth}
      \centering
      \includegraphics[width=\linewidth]{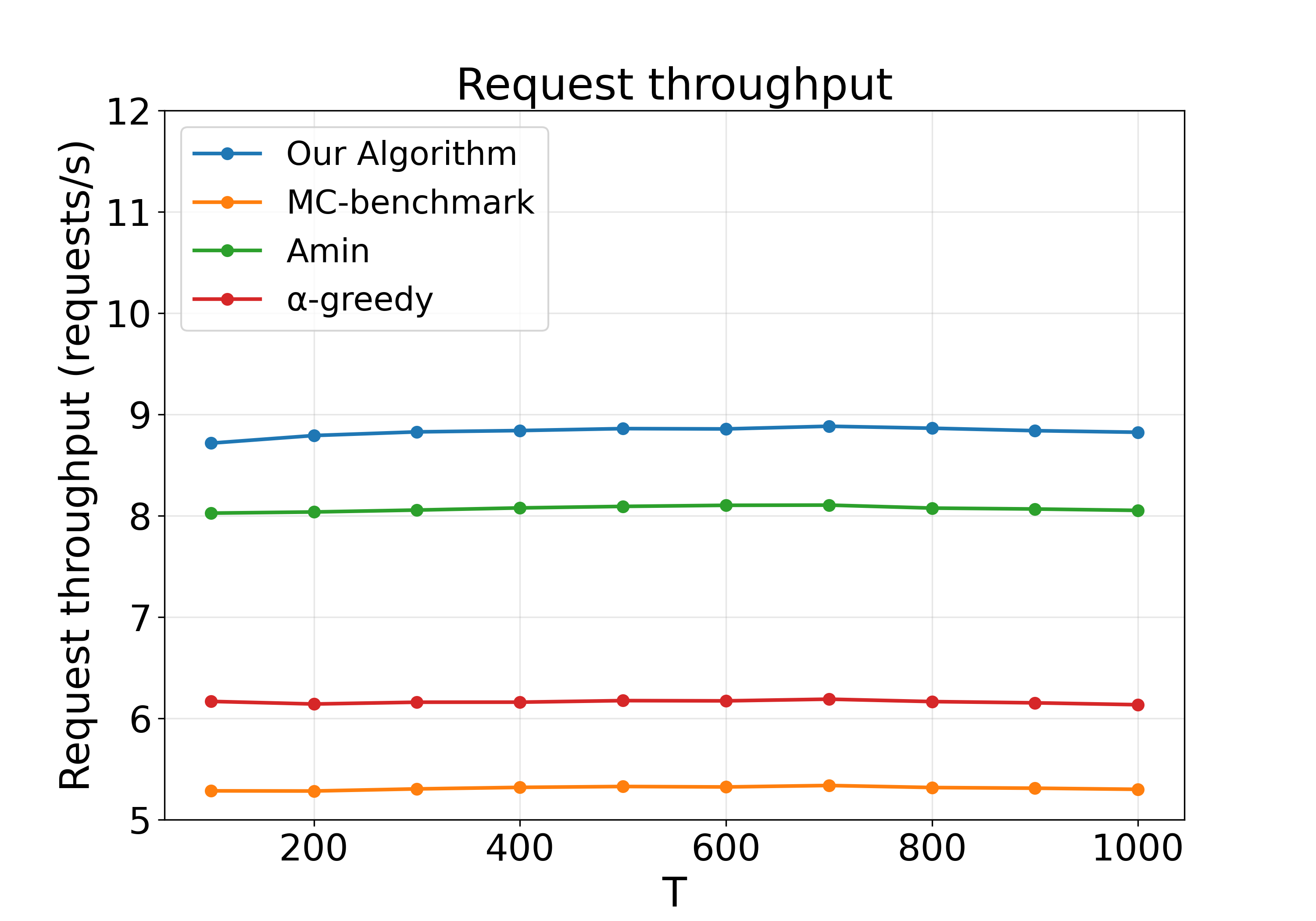} 
    \end{minipage}
    \begin{minipage}[t]{0.24\linewidth}
      \centering
      \includegraphics[width=\linewidth]{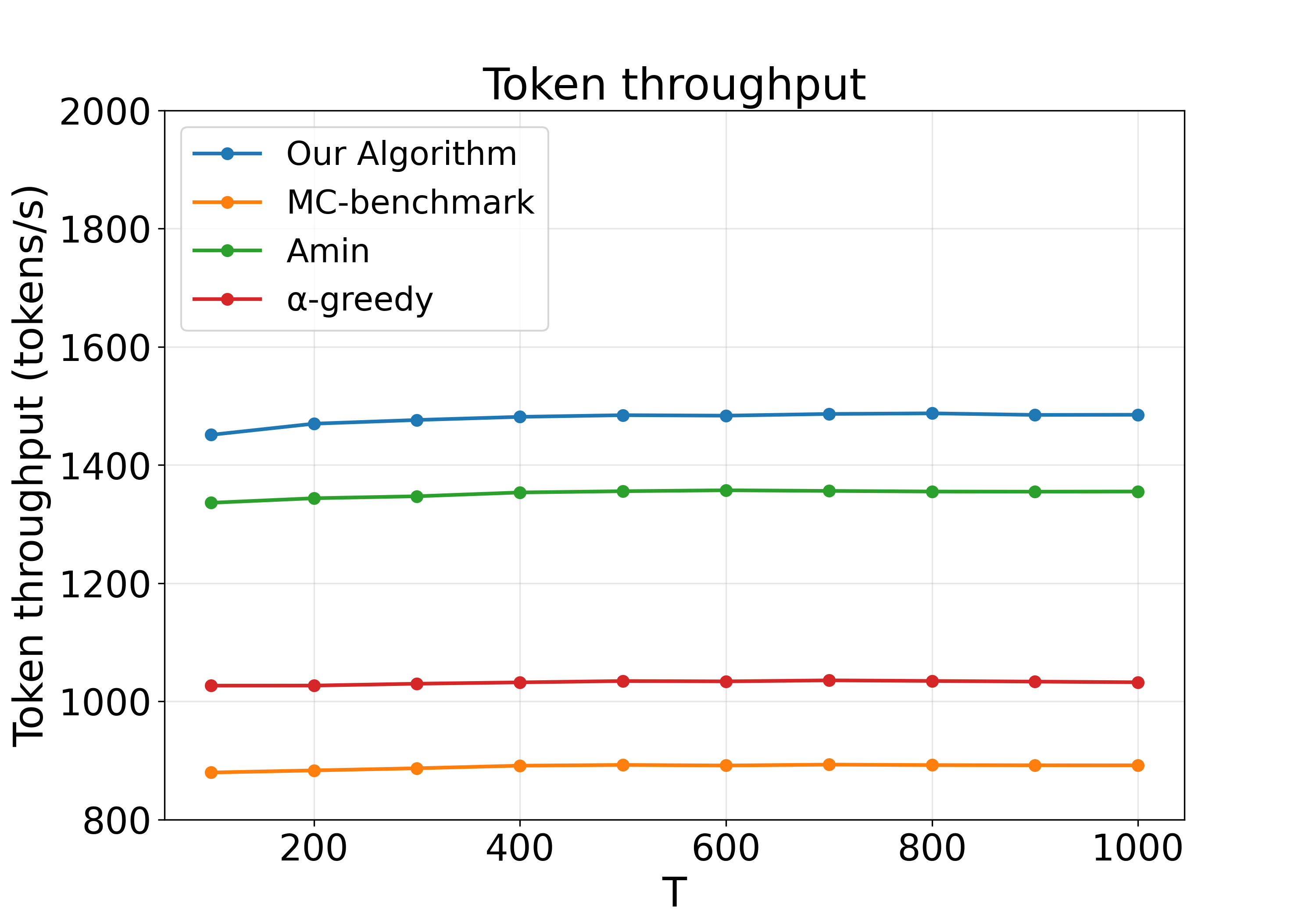} 
    \end{minipage}
  \end{minipage}

\caption{(Up) Performance metrics across scheduling algorithms under the low demand setting. (Down) Performance metrics across scheduling algorithms under the high demand setting. The performance metrics (from left to right) are average latency, 95\% latency, request throughput, and token throughput.}\label{f2}
\end{figure}

As observed in Figure \ref{f2}, our algorithm shows competitive performance with the benchmarks, which demonstrates the benefit of controlling the rate that requests are activated. By controlling activation rates, our policy avoids overly aggressive admissions that can trigger overflows, leading to frequent evictions and waste. We also observe that the performance gap narrows compared to the synthetic data. A plausible explanation is that the workload of real-world datasets is larger (i.e., the prefill and output lengths are longer than the synthetic data), which requires a smaller activation budget $b$. This induces larger fluctuations in instantaneous memory occupancy, causing increased variability and diminishing the advantage of activation flow control.

\section{Discussion}
In this paper, we study the LLM serving system with stochastic and heterogeneous request arrivals. Motivated by the results in queueing theory, we propose an easy-to-implement flow-controlled scheduling algorithm, which limits the rate that requests are activated. We identify a sufficient condition under which no scheduling policy can achieve stability and prove that our algorithm can achieve low overflow probability under certain criteria. Numerical experiments on both synthetic and practical requests indicate that our algorithm achieve competitive throughput and latency relative to the benchmarks. These results suggest that controlling the activation rate can be beneficial for the inference serving system.

There are several directions that merit future work. Firstly, incorporating the output length predictions \cite{zheng2023response} with LLM serving can potentially improve the performance of the serving system and allows for more informative decision-making. A second direction is to design an adaptive scheduling policy that combines both our flow-control mechanism with aggressive greedy heuristics. While our numerical results suggest that a greedy policy can frequently overflow in high-traffic regimes, it may outperform our algorithm under low-traffic scenarios because our activation mechanism is conservative. This motivates dynamic policies that switch the mechanism according to the traffic and real-time system conditions. Finally, extending our framework to multi-turn interaction systems presents an intriguing direction. In practice, consecutive prompts within a chat are correlated and share context. This allows prompt caching to become a critical strategy for improving the efficiency of the LLM inference. To more accurately reflect real-world deployments, future work could study the inference system with multi-turn requests. In addition, our discrete-time model does not explicitly account for the computational cost of the prefill stage, where the entire input prompt is processed in one (or a few chunked) forward passes. Incorporating prefill computation time into the theoretical framework could yield tighter performance characterizations.

\bibliographystyle{plainnat}
\bibliography{reference}

\appendix

\section{Proofs}\label{app1}
In this section, we provide proofs of technical results in the main text that are omitted.

\subsection{Proof of Proposition \ref{prop1}}\label{app_a1}
For notational convenience, we define 
\begin{equation*}
    N(T):=\sum_{t=1}^{T} n_t
\end{equation*}
as the overall number of requests arrived during the time period $\{1,2,\ldots,T\}$, and $N_k(T)$ denotes the number of type $k$ requests arrived during the same period. For each request of type $k$, we note that the minimal KV cache workload to complete would be
\begin{equation*}
    w_k=\sum_{j=1}^{o^{(k)}} (l^{(k)}+ j)=l^{(k)}o^{(k)} + \frac{o^{(k)}(o^{(k)}+1)}{2}=l^{(k)}o^{(k)} + \frac{1}{2}\left(o^{(k)} + (o^{(k)})^2\right).
\end{equation*}
As a result, the (minimum) number of tokens required to complete these requests would be:
\begin{equation*}
    \mathcal{N}(T) = \sum_{k=1}^{m} N_k(T) \left( l^{(k)}o^{(k)} + \frac{1}{2} \left(o^{(k)} + (o^{(k)})^2 \right) \right),
\end{equation*}
with expectation:
\begin{align*}
    \mathbb E \mathcal{N}(T) &= \mathbb E\left[\sum_{k=1}^{m} N_k(T) \left( l^{(k)}o^{(k)} + \frac{1}{2} \left(o^{(k)} + (o^{(k)})^2 \right) \right)\right] \\&= T\sum_{k=1}^{m}\lambda_k \left(l^{(k)}o^{(k)} + \frac{1}{2} \left(o^{(k)} + (o^{(k)})^2\right) \right) \\&= T\sum_{k=1}^{m}\lambda_k w_k.
\end{align*}

Since the maximum KV cache budget is $M$. The system can only utilize up to $M$ tokens at each time step for prefilling or decoding. Thus, the cumulative available KV cache budget is at most $MT$. Let $C_k(T)$ be the number of type $k$ requests completed by the end of $T$, and let $C(T)=\sum_{k=1}^m C_k(T)$. Therefore, summing all the completed requests by time $T$ yields the following inequality:
\begin{equation*}
    \sum_{k=1}^{m} C_k(T) w_k \leq MT.
\end{equation*}
Define the additional tokens required to complete all requests arrived during the time period $\{1,2,\ldots,T\}$ would be as follows.
\begin{equation}
    R(T) = \sum_{k=1}^{m} \left(N_k(T) - C_k(T)\right)w_k
\end{equation}
Taking the expectation on both sides yield
\begin{equation*}
    \mathbb E R(T) =\mathbb E \left[  \sum_{k=1}^{m} \left(N_k(T) - C_k(T)\right)w_k \right] \geq T\left(\sum_{k=1}^{m}\lambda_k w_k - M\right) = \delta T,
\end{equation*}
where we define $\delta = \sum_{k=1}^{m}\lambda_k w_k - M>0$. 

Next, we define $w_{\max}:=\max_{k\in[m]} w_k < \infty$. Define the total number of unfinished
arrived requests by time $T$ as
\[Q(T) ~:=~ \sum_{k=1}^{m} \left(N_k(T)-C_k(T)\right).\]
Since each unfinished request contributes at most $w_{\max}$ to $R(T)$, we have $R(T)\le w_{\max} Q(T)$,
and therefore
\[\mathbb{E}[Q(T)] \geq \frac{\mathbb{E}[R(T)]}{w_{\max}} \geq \frac{\delta T}{w_{\max}}. \]
By the laws of large numbers, as $T$ goes to infinity,
\begin{equation*}
    \frac{N_k(T)}{T}\rightarrow \lambda_k
\end{equation*}
holds almost surely, and thus
\begin{equation*}
    \frac{R(T)}{T} \geq \frac{\sum_{k=1}^{m} N_k(T)w_k-MT}{T}\rightarrow \frac{\sum_{k=1}^{m} \lambda_k w_k}{T} - M = \delta>0.
\end{equation*}
Recall that $R(T)\le w_{\max} Q(T)$, we have
\begin{equation*}
    \frac{Q(T)}{T} \geq \frac{\delta}{w_{\max}}>0
\end{equation*}
holds almost surely. This completes the proof because we can see that the number of unfinished requests would explode over time.

\subsection{Proof of Theorem \ref{thm1}}

Under Algorithm~\ref{alg1}, at each slot $t$, the system activates at most $b_k$ waiting requests of each type $k$ and then processes all active requests and removes completed ones. Therefore, the memory usage (at any $t$) would not exceed
\begin{equation}
    \sum_{k=1}^m \sum_{j=1}^{o^{(k)}} b_k (l^{(k)}+ j) = \sum_{k=1}^m b_k \left( l^{(k)}o^{(k)} + \frac{1}{2} \left(o^{(k)} + (o^{(k)})^2 \right) \right) <M,
\end{equation}
meaning that the KV cache constraint is always satisfied. That is, no eviction would ever happen. It then suffices to show that the system would be stable under this algorithm.

Let $Q_{t,k}$ denote the number of type $k$ requests in the waiting queue at the end of slot $t$, $n_{i,t}$ is the number of newly arrived requests of type $k$. Under Algorithm~\ref{alg1}, the system admits
\[
    a_{t,k} := \min\{b_k,\, Q_{t-1,k} + n_{t,k}\}
\]
type $k$ requests at slot $t$, hence the waiting queue evolves as
\begin{equation*}
    Q_{t,k}
    = Q_{t-1,k} + n_{t,k} - a_{t,k}
    = \bigl(Q_{t-1,k} + n_{t,k} - b_k\bigr)^+.
\end{equation*}
Since Algorithm~\ref{alg1} manages each request type $k$ independently using a dedicated budget $b_k$, the dynamics of the waiting queue $Q_{t,k}$ for each type $k$ are decoupled from other types. We can consider them separately.

For the type $k$ requests, consider the following quadratic Lyapunov function: $V(Q_{t,k}) = \frac{1}{2} Q_{t,k}^2$. Then we can bound the drift $\Delta(Q_{t,k}) = \mathbb{E}[V(Q_{t+1,k}) - V(Q_{t,k}) \mid Q_{t,k}]$ as follows:
\begin{align}\label{eq5}
    \Delta(Q_{t,k}) &= \frac{1}{2} \mathbb{E} \left[ \left( (Q_{t-1,k} + n_{t,k} - b_k)^+ \right)^2 - Q_{t-1,k}^2 \mid Q_{t-1,k} \right] \notag \\
    &\le \frac{1}{2} \mathbb{E} \left[ (Q_{t-1,k} + n_{t,k} - b_k)^2 - Q_{t-1,k}^2 \mid Q_{t-1,k} \right] \notag \\
    &= \mathbb{E} \left[ Q_{t-1,k}(n_{t,k} - b_k) + \frac{1}{2}(n_{t,k} - b_k)^2 \mid Q_{t-1,k} \right].
\end{align}
Let $\epsilon_k = b_k - \lambda_k>0$. Then the first term in Inequality \eqref{eq5} simplifies to:
\[
    \mathbb{E} [Q_{t-1,k}(n_{t,k} - b_k) \mid Q_{t-1,k}] = Q_{t-1,k}(\lambda_k - b_k) = -\epsilon_k Q_{t-1,k}.
\]
Note that the arrival process has a bounded second moment, i.e., $\mathbb{E}[(n_{t,k} - b_k)^2] \le B_k < \infty$. The drift inequality becomes:
\begin{equation}
    \Delta(Q_{t,k}) \le -\epsilon_k Q_{t-1,k} + \frac{B_k}{2}.
\end{equation}
When the queue length is sufficiently large (e.g., $Q_{t-1,k} > B_k / \epsilon_k$), the drift is strictly negative. 
According to the Foster-Lyapunov criterion (Theorem 11.3.4 in \citep{meyn2012markov}), we can see that the queue $Q_{t,k}$ is positive recurrent for all $k$. Hence, the entire system is stable.

\subsection{Proof of Proposition \ref{prop2}}
We follow the same notations we used in Appendix \ref{app_a1}. We index the requests by $p_1 ,p_2,\ldots$ according to their order of arrival. Given the number of requests arrived by time $T$, $N(T)$, we are able to obtain the following equation:
\begin{align*}
    \mathcal N(T) = \sum_{i=1}^{N(T)} \left(l_io_i + \frac{1}{2}\left(o_i+o_i^2\right)\right),
\end{align*}
where $\mathcal N(T)$ is the minimum number of tokens required to complete these requests. Taking the expectation on both sides yield the following
\begin{align*}
    \mathbb E \mathcal{N}(T) &=  \mathbb E \left[\sum_{i=1}^{N(T)} \left(l_io_i + \frac{1}{2}\left(o_i+o_i^2\right)\right)\right]\\
    &=\mathbb E N(T) \mathbb E \left[\left(lo + \frac{1}{2}\left(o+o^2\right)\right)\right]\\
    & = T\lambda \mathbb E \left[\left(lo + \frac{1}{2}\left(o+o^2\right)\right)\right].
\end{align*}
Here $(l,o)\sim \nu$. Next, we define $\delta = \lambda \mathbb E \left[\left(lo + \frac{1}{2}\left(o+o^2\right)\right)\right] - M >0$. Let $w_{\max} = \frac{3}{2}C^2+\frac{1}{2}C$. This gives us
\begin{align*}
    \left(lo + \frac{1}{2}\left(o+o^2\right)\right) \leq \frac{3}{2}C^2+\frac{1}{2}C = w_{\max}.
\end{align*}
As a result, following the same proof technique we used in Appendix \ref{app_a1}, we find that
\begin{align*}
    \frac{Q(T)}{T}\geq \frac{\delta}{w_{\max}}
\end{align*}
and completes the proof.

\subsection{Proof of Theorem \ref{thm2}}
Following Algorithm \ref{alg2}, we can see that the number of newly activated requests is up to $B_t$. For any $s\in \mathbb N$, define the number of requests that join the active set as $a_{s}$ at slot $s$ ($a_{s}\leq B_s$). Without loss of generality, let $a_s =0$ for any $s\leq 0$. Note that any activated requests will keep decoding unless being evicted or completed. As a result, the overall KV cache usage at time $t$ would be
\begin{align*}
    U_t &= \sum_{s=1}^{t}\sum_{p=1}^{a_s} \left( l_{s,p}+(t-s+1) \right)\mathbbm 1\left(o_{s,p}\geq t-s+1 \right)\\
    & = \sum_{s=\max\{1,t-C+1\}}^{t}\sum_{p=1}^{a_s} \left( l_{s,p}+(t-s+1) \right)\mathbbm 1\left(o_{s,p}\geq t-s+1 \right)\\
    & = \sum_{q=1}^{C}\sum_{p=1}^{a_{t-q+1}} \left( l_{t-q+1,p}+q \right)\mathbbm 1\left(o_{t-q+1,p}\geq q \right),
\end{align*}
where we define $l_{s,p}$ as the $p^{th}$ requests that join the active set at time $s$. Recall that $a_{s}\leq B_s$, we obtain the following:
\begin{align*}
    U_t& = \sum_{q=1}^{C}\sum_{p=1}^{a_{t-q+1}} \left( l_{t-q+1,p}+q \right)\mathbbm 1\left(o_{t-q+1,p}\geq q \right) \\
    &\leq \sum_{q=1}^{C}\sum_{p=1}^{B_{t-q+1}} \left( l_{t-q+1,p}+q \right)\mathbbm 1\left(o_{t-q+1,p}\geq q \right) : = \hat U_t.
\end{align*}
In the above inequality, we (potentially) define some fictitious requests: for any $a_s<B_s$, we add the active set with $(B_s-a_s)$ fictitious requests. These requests are also i.i.d. sampled from the distribution $\nu$. Taking the expectation on the right hand side yields:
\begin{align*}
    \mathbb E[\hat U_t] &= \mathbb E \left[ \sum_{q=1}^{C}\sum_{p=1}^{B_{t-q+1}} \left( l_{t-q+1,p}+q \right)\mathbbm 1\left(o_{t-q+1,p}\geq q \right) \right]\\
    & = \mathbb E[B_t] \mathbb E_{(l,o)\sim \nu} \left[ \sum_{q=1}^{C} \left( l+q \right)\mathbbm 1\left(o\geq q \right) \right]\\
    & = b \mathbb E_{(l,o)\sim \nu} \left[ \sum_{q=1}^{o} \left( l+q \right) \right]\\
    & = b\mathbb E_{(l,o)\sim \nu} \left[ lo + \frac{1}{2}(o+o^2) \right]\leq (1-\epsilon)M.
\end{align*}

Next, note that $U_t\leq \hat U_t$, we attempt to bound $\mathbb{P}(U_t>M)$ by its upper bound $\mathbb{P}(\hat U_t>M)$:
\begin{align*}
    \mathbb{P}(U_t>M)\le \mathbb{P}(\hat U_t>M) = \mathbb{P}\big(\hat U_t-\mathbb{E}[\hat U_t] \ge M-\mathbb{E}[\hat U_t]\big) \le \mathbb{P}\big(\hat U_t-\mathbb{E}[\hat U_t] \ge \epsilon M\big).
\end{align*}
By Assumption \ref{as2},
\[0\leq \left( l_{t-q+1,p}+q \right)\mathbbm 1\left(o_{t-q+1,p}\geq q \right)\leq  l_{t-q+1,p}+q  \leq 2C,\]
and
\[0\leq \sum_{p=1}^{B_{t-q+1}} \left( l_{t-q+1,p}+q \right)\mathbbm 1\left(o_{t-q+1,p}\geq q \right)\leq 2AC.\]
Then McDiarmid's inequality yields
\[
\mathbb{P}\big(\hat U_t-\mathbb{E}[\hat U_t] \ge \epsilon M\big) \le \exp\!\left( -\frac{2(\epsilon M)^2}{C(2AC)^2+AC(2C)^2} \right) = \exp\!\left( -\frac{\epsilon^2 M^2}{2(A^2+A)C^3} \right).
\]
Consequently, for all $t$,
\begin{align}\label{eq10}
    \mathbb{P}(U_t>M)\le \mathbb{P}(\hat U_t>M)\le \exp\!\left(-C(A,\epsilon)M^2\right).
\end{align}
Here, $C(A,\epsilon)=\frac{\epsilon^2}{2(A^2+A)C^3}>0$. Putting the above together,
\[
\mathbb{E}\Big[\sum_{t=1}^T I_t\Big] \le \sum_{t=1}^T \mathbb{P}(U_t>M) \le \sum_{t=1}^T \mathbb{P}(\hat U_t>M) \le T\exp\!\left(-C(A,\epsilon)M^2\right).
\]
This completes the proof.

\section{Additional Numerical Results}
\begin{figure}
  \centering

  \begin{minipage}{1\textwidth}
    \centering
    \begin{minipage}[t]{0.48\linewidth}
      \centering
      \includegraphics[width=\linewidth]{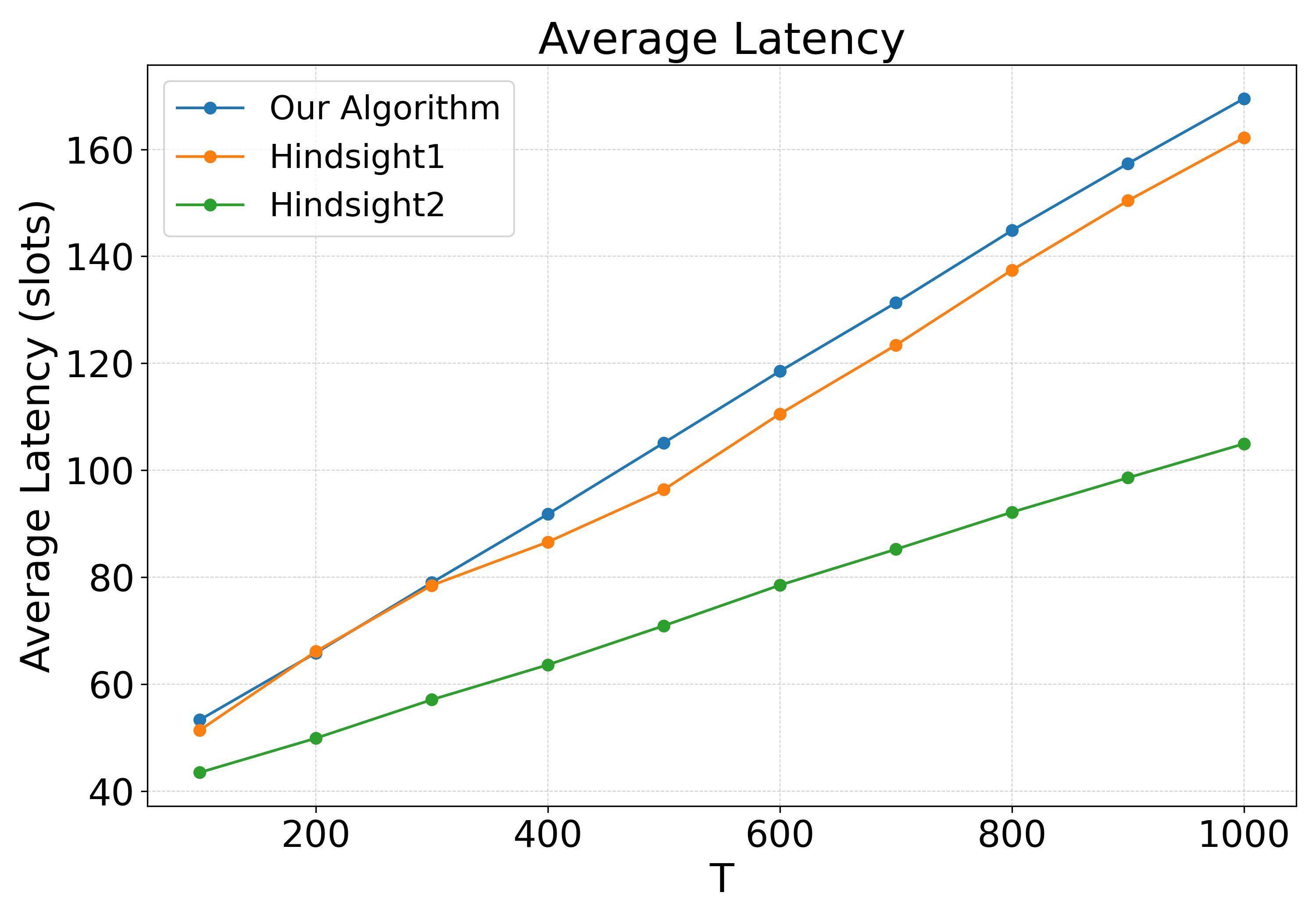} 
    \end{minipage}\hfill
    \begin{minipage}[t]{0.48\linewidth}
      \centering
      \includegraphics[width=\linewidth]{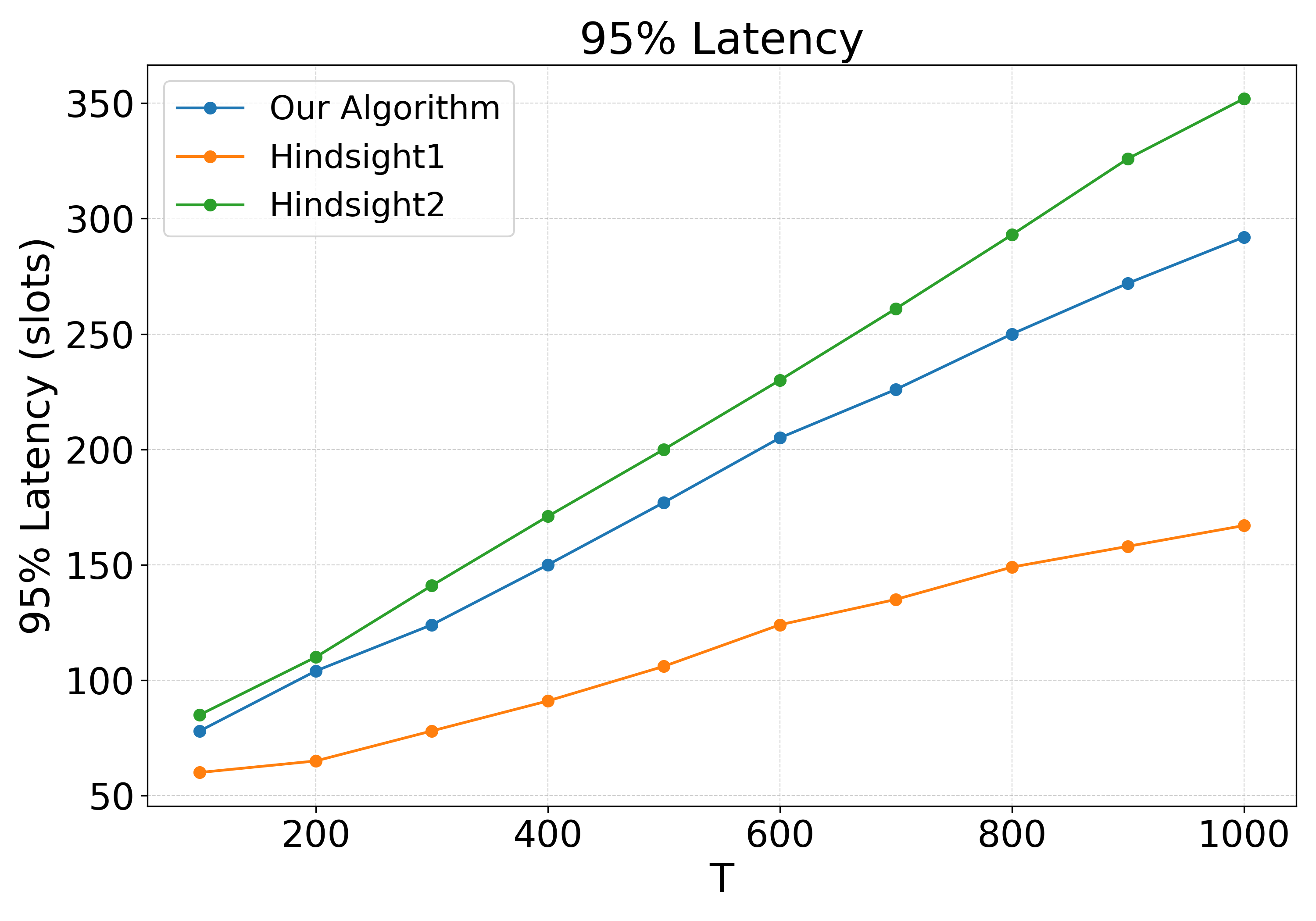} 
    \end{minipage}\hfill
  \end{minipage}

  \begin{minipage}{1\textwidth}
    \centering
    \begin{minipage}[t]{0.48\linewidth}
      \centering
      \includegraphics[width=\linewidth]{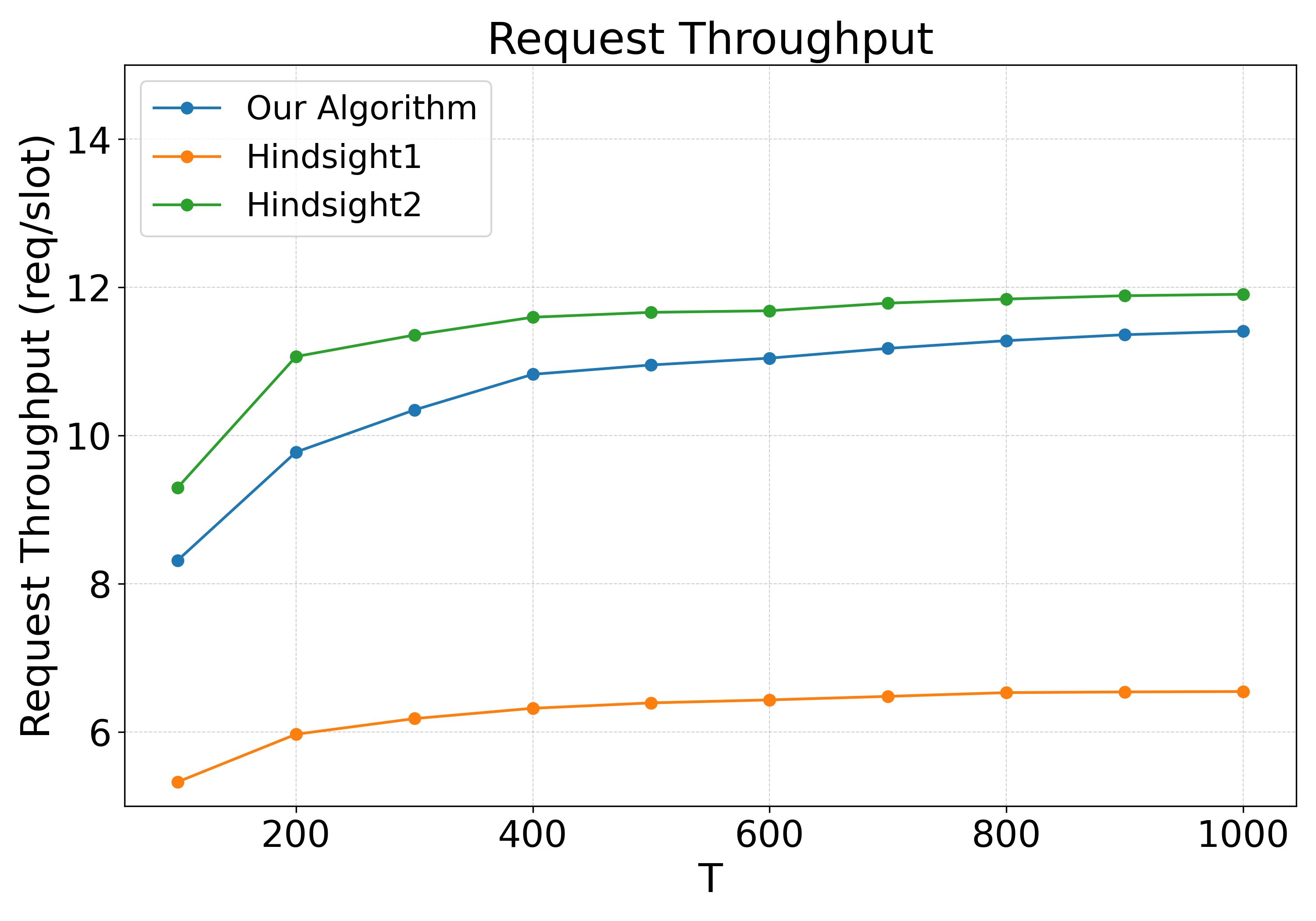} 
    \end{minipage}\hfill
    \begin{minipage}[t]{0.48\linewidth}
      \centering
      \includegraphics[width=\linewidth]{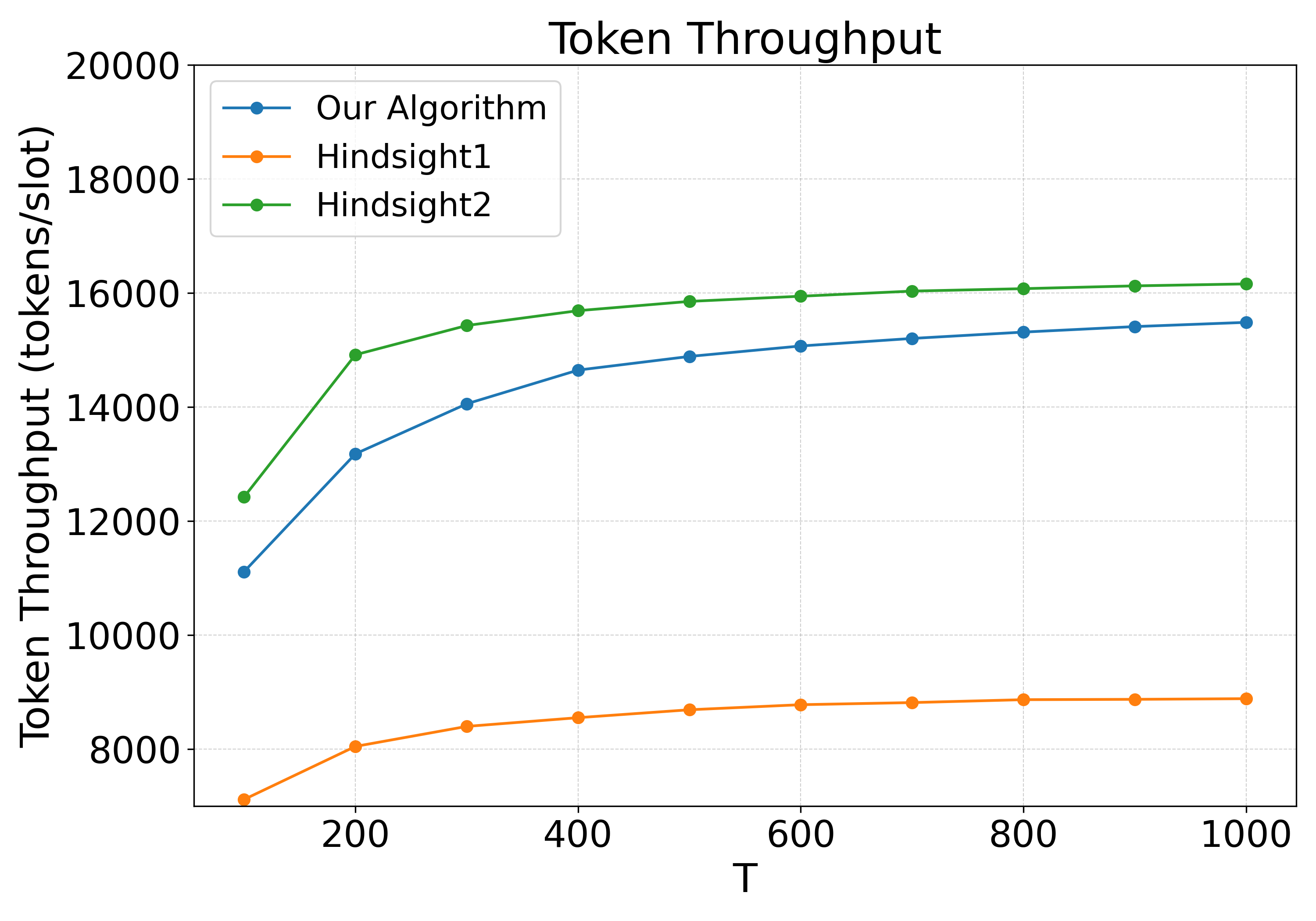} 
    \end{minipage}\hfill

  \end{minipage}

\caption{Performance of our scheduling algorithm against the two benchmarks. (Up) The performance metrics (from left to right) are average latency and 95\% latency. (Down) The performance metrics (from left to right) are request throughput and token throughput.}\label{f3}
\end{figure}

We further compare our algorithm against two hindsight solutions: \emph{hindsight1}, which minimizes tail latency, and \emph{hindsight2}, which minimizes other performance metrics. Notably, these two hindsight solutions themselves yield different optimal decisions. The results are presented in Figure \ref{f3}.

\end{document}